\documentclass[11pt]{article}

\usepackage[preprint]{acl}
\usepackage{times}
\usepackage{latexsym}
\usepackage[T1]{fontenc}
\usepackage[utf8]{inputenc}
\usepackage{microtype}
\usepackage{inconsolata}
\usepackage{graphicx}
\usepackage{amssymb}
\usepackage{amsmath}
\usepackage{booktabs}
\usepackage{tabularx}
\usepackage{makecell}
\usepackage[table]{xcolor}
\usepackage{multirow}
\newcolumntype{L}[1]{>{\raggedright\arraybackslash}p{#1}}
\newcolumntype{C}[1]{>{\centering\arraybackslash}p{#1}}
\newcolumntype{Y}{>{\centering\arraybackslash}X}

\title{ElasticFlow: One-Step Physics-Consistent Policy with Elastic Time Horizons for Language-Guided Manipulation}

\author{
	\textbf{Kewei Chen}\textsuperscript{\rm 1, 2},
	\textbf{Yayu Long}\textsuperscript{\rm 1, 2},
	\textbf{Shuai Li}\textsuperscript{\rm 3},
	\textbf{Mingsheng Shang}\textsuperscript{\rm 1, 2}\thanks{Corresponding author.} \\
	\textsuperscript{\rm 1}Chongqing Institute of Green and Intelligent Technology, Chinese Academy of Sciences\\
	\textsuperscript{\rm 2}Chongqing School, University of Chinese Academy of Sciences\\
	\textsuperscript{\rm 3}Faculty of Information Technology and Electrical Engineering, University of Oulu, Finland\\
	\{chenkewei24, longyayu24\}@mails.ucas.ac.cn, shuai.li@oulu.fi, msshang@cigit.ac.cn
}

\begin{document}
\maketitle
\begin{abstract}

	Diffusion policies have demonstrated exceptional performance in embodied AI. However, their iterative denoising process results in high latency, and existing acceleration methods often sacrifice physical consistency. To address this, we propose ElasticFlow, a distillation-free, physics-consistent one-step policy framework. We reconstruct the Mean Field Theory by directly modeling the average velocity field, enabling a direct single-step mapping from noise to action. Addressing the Temporal Heterogeneity of robotic tasks, we introduce the Elastic Time Horizons mechanism. This mechanism effectively overcomes Spectral Bias by explicitly encoding control granularity, achieving efficient alignment between semantic instructions and physical execution horizons. Experiments on benchmarks such as LIBERO, CALVIN, and RoboTwin demonstrate that ElasticFlow achieves efficient 1-NFE inference (approximately 71Hz). Furthermore, it outperforms state-of-the-art methods, including OpenVLA and $\pi_0$, on long-horizon tasks, highlighting its potential for efficient, robust, and semantically aligned control.

\end{abstract}

\section{Introduction}
\label{sec:intro}

Constructing Generalist Policies that map visual observations and \textbf{natural language instructions} to continuous action spaces is a pivotal objective in Embodied AI~\citep{brohan2023rt2, team2024octo, black2024pi_0}. While Diffusion Policies have become a dominant paradigm due to their robust multi-modal modeling~\citep{chi2023diffusion}, their reliance on iterative denoising necessitates dozens of network Function Evaluations (NFEs)~\citep{songdenoising}. This results in high inference latency, constraining the robot's real-time responsiveness to dynamic environments, such as capturing moving objects~\citep{prasad2024consistency}.

To overcome this speed bottleneck, existing solutions primarily focus on acceleration strategies like Consistency Models or progressive distillation~\citep{song2023consistency, salimans2022progressive}. However, these methods often necessitate complex training pipelines and are prone to instability or insufficient mode coverage. Critically, mere acceleration frequently neglects geometric trajectory characteristics, leading to actions that lack Physics-Consistency. Similarly, auto-regressive VLA models typically rely on discretized token prediction or low-frequency diffusion heads, limiting the balance between high-frequency control and physical continuity~\citep{kim2025openvla, wen2025diffusionvla}.

To address these limitations, we propose \textbf{ElasticFlow}, a physics-consistent one-step policy framework. Inspired by MeanFlow~\citep{geng2025mean}, we reconstruct the Mean Field Theory by directly modeling the \textbf{Average Velocity Field} rather than the instantaneous velocity field. Utilizing the Fundamental Theorem of Calculus, we derive the MeanFlow Identity, a differential constraint that enables learning a direct mapping from noise to data. This formulation mathematically eliminates multi-step integration and implicitly incorporates curvature correction to ensure trajectory smoothness.

Addressing the temporal heterogeneity of robotic tasks, we further introduce the \textbf{Elastic Time Horizons} mechanism. Traditional fixed-horizon methods suffer from Spectral Bias when handling mixed-frequency signals. By explicitly encoding the time span via parameters $(r, t)$, ElasticFlow dynamically aligns semantic instructions with physical execution horizons. It can generate millisecond-level transient control quantities and plan macroscopic trajectories spanning seconds within a unified weight space. Our main contributions are summarized as follows:

\textbf{(1) ElasticFlow Policy Framework:} We formulate robot action generation as an average velocity field learning problem. By leveraging the MeanFlow Identity, we achieve a distillation-free, 1-NFE policy that ensures physical smoothness without complex training pipelines.

\textbf{(2) Elastic Time Abstraction Mechanism:} We propose a dual-parameter time encoding strategy that endows the model with explicit perception of the control horizon. This enables a single network to adaptively handle diverse task requirements and mitigates cumulative errors in long-range tasks.

\textbf{(3) SOTA Performance and Efficiency:} Experiments on benchmarks including LIBERO, CALVIN, and RoboTwin demonstrate that ElasticFlow achieves an inference frequency of approximately 71Hz. It outperforms state-of-the-art methods, such as OpenVLA and $\pi_0$, on long-horizon tasks (e.g., 95.6\% on LIBERO-Long).

\section{Related Work}
\label{sec:related_work}

\paragraph{Generative Policies in Robot Learning}
Generative models, such as Diffusion Policy~\citep{chi2023diffusion, ze20243d} and ACT~\citep{zhao2023learning}, have represented the state-of-the-art in robotic manipulation. While Diffusion Policy effectively captures multi-modal action distributions, it is inherently limited by the high latency of iterative denoising solvers~\citep{chi2023diffusion}. Conversely, ACT utilizes action chunking to mitigate inference pressure but typically relies on heuristic temporal ensembling, which often leads to over-smoothing artifacts and a lack of strict physical consistency~\citep{zhao2023learning, prasad2024consistency}. 

To address the sampling bottleneck, acceleration techniques like Consistency Models~\citep{song2023consistency} and Progressive Distillation~\citep{salimans2022progressive} have been explored; however, they often necessitate complex teacher-student training pipelines or compromise trajectory smoothness. Although recent Flow Matching methods~\citep{lipman2023flow, black2024pi_0, zhong2025flowvla, lin2025hifvla} offer a more direct generation path, most still fundamentally rely on iterative ODE integration. In contrast, ElasticFlow achieves physically consistent, distillation-free 1-NFE generation by extending the average velocity field theory~\citep{geng2025mean} to dynamic control. By incorporating a geometric curvature correction term, ElasticFlow bypasses both the sampling latency of diffusion models and the averaging artifacts of ACT.

\paragraph{VLA Models and Long-Horizon Planning}
Vision-Language-Action (VLA) models typically employ auto-regressive token prediction~\citep{brohan2023rt2, kim2025openvla}, which introduces quantization errors and latency. To mitigate forgetting in long-horizon tasks, recent approaches incorporate external memory~\citep{shi2025memoryvla} or chain-of-thought reasoning~\citep{zhao2025cot}, increasing architectural complexity. Hybrid models like DiffusionVLA~\citep{wen2025diffusionvla} improve generalization but remain constrained by the latency of iterative denoising. In contrast, ElasticFlow introduces \textbf{Elastic Time Horizons} to efficiently unify high-frequency reactive control and long-range planning within a single network.

\section{Method}

\label{sec:method}

This section details \textbf{ElasticFlow}, a Physics-Consistent one-step policy learning framework for embodied AI. This framework aims to mitigate the contradiction between inference speed and physical constraints in traditional diffusion policies. By establishing an \textbf{Average Velocity Field} between language instructions and continuous physical actions, it achieves high-frequency action generation supporting \textbf{Elastic Time Horizons} under a unified mathematical form.

\subsection{Problem Definition: From Local Tangent to Global Average Field}

We model language-conditioned robot manipulation tasks as a flow transformation process from a prior noise distribution $p_0$ to a target action manifold $p_1$. Let $o \in \mathcal{O}$ be the robot's multi-modal observation sequence, and $\ell \in \mathcal{L}$ be the natural language instruction. The action space is defined as $x \in \mathbb{R}^{T_h \times D}$, representing an Action Chunk of length $T_h$, where $D$ is the action dimension. The goal is to learn a policy $\pi_\theta(x|o, \ell)$ capable of generating action trajectories that satisfy physical dynamics constraints and are semantically aligned with extremely low latency.

Traditional diffusion-based methods view action generation as an iterative Markov denoising process, limiting policy inference frequency (typically $<20$Hz). In contrast, we reconstruct action generation as a Flow Matching problem. Traditional flow matching drives Ordinary Differential Equations (ODEs) by regressing the Instantaneous Velocity Field $v(z_\tau, \tau)$. However, instantaneous velocity is essentially the local tangential derivative of the trajectory. It lacks a direct description of global trajectory geometry, necessitating multiple function evaluations (NFE) via numerical integration during inference.

To achieve one-step generation, we introduce the \textbf{Average Velocity Field} $u(z_t, r, t)$. Unlike instantaneous velocity, average velocity is defined as the normalized displacement of the flow within the elastic time interval $[r, t]$:

\begin{equation}
	u(z_t, r, t) \triangleq \frac{1}{t-r} \int_{r}^{t} v(z_\tau, \tau) d\tau
\end{equation}

Here, $z_t \in \mathbb{R}^D$ is the state at time $t$, and $v(z_\tau, \tau)$ represents the instantaneous velocity field of the flow at time $\tau$. This integral path is defined along the Characteristic Line, directly associating the current state with historical states. Its core advantage is that if the average velocity from $r=0$ (noise domain) to $t=1$ (data domain) can be accurately predicted, the target action satisfying physical consistency can be directly recovered via the single-step mapping $\hat{x} = z_1 - u(z_1, 0, 1)$, without iterative solving.

\subsection{Policy Learning based on ElasticFlow Identity}

To learn the aforementioned average velocity field, we construct supervision signals using the ElasticFlow Identity. Based on the Fundamental Theorem of Calculus, we derive the following intrinsic differential constraint between average velocity $u$ and instantaneous velocity $v$ (detailed derivation proof in Appendix \ref{app:derivation}):

\begin{equation}
	u(z_t, r, t) = v(z_t, t) - (t-r)\frac{d}{dt}u(z_t, r, t)
	\label{eq:ElasticFlow_identity}
\end{equation}

Here, $\frac{d}{dt}$ denotes the Total Derivative with respect to time $t$. This identity reveals the Intrinsic Geometric Constraint between average velocity $u$ and instantaneous velocity $v$.

Under the Conditional Flow Matching paradigm, we construct the optimal transport path, where the ground truth instantaneous velocity is $v(z_t, t) = x_{target} - x_{noise}$.

Equation (\ref{eq:ElasticFlow_identity}) provides a clear physical interpretation: $u(z_t, r, t)$ represents the "effective average transport velocity" backtracking from the current time $t$ to the start time $r$ under semantic instruction constraints. The second term on the right side, $(t-r)\frac{d}{dt}u$, acts as a \textbf{Manifold Curvature Correction Term}. As the time span $t-r$ increases, this term corrects trajectory deviations caused by the non-linear manifold. By explicitly modeling this physical quantity, the policy network is forced to learn the global geometric features of the action trajectory rather than local gradients. This inherently suppresses high-frequency jitter in action sequences and improves control smoothness.

By explicitly modeling this physical quantity, the policy network is compelled to learn global geometric features of action trajectories instead of local gradients. This global perspective modeling not only guarantees the precision of one-step generation but also inherently suppresses high-frequency jitter in action sequences, enhancing control smoothness.

\subsection{Elastic Time Abstraction and Network Architecture}

\begin{figure*}[t]
	\centering
	\footnotesize
	\includegraphics[width=\textwidth]{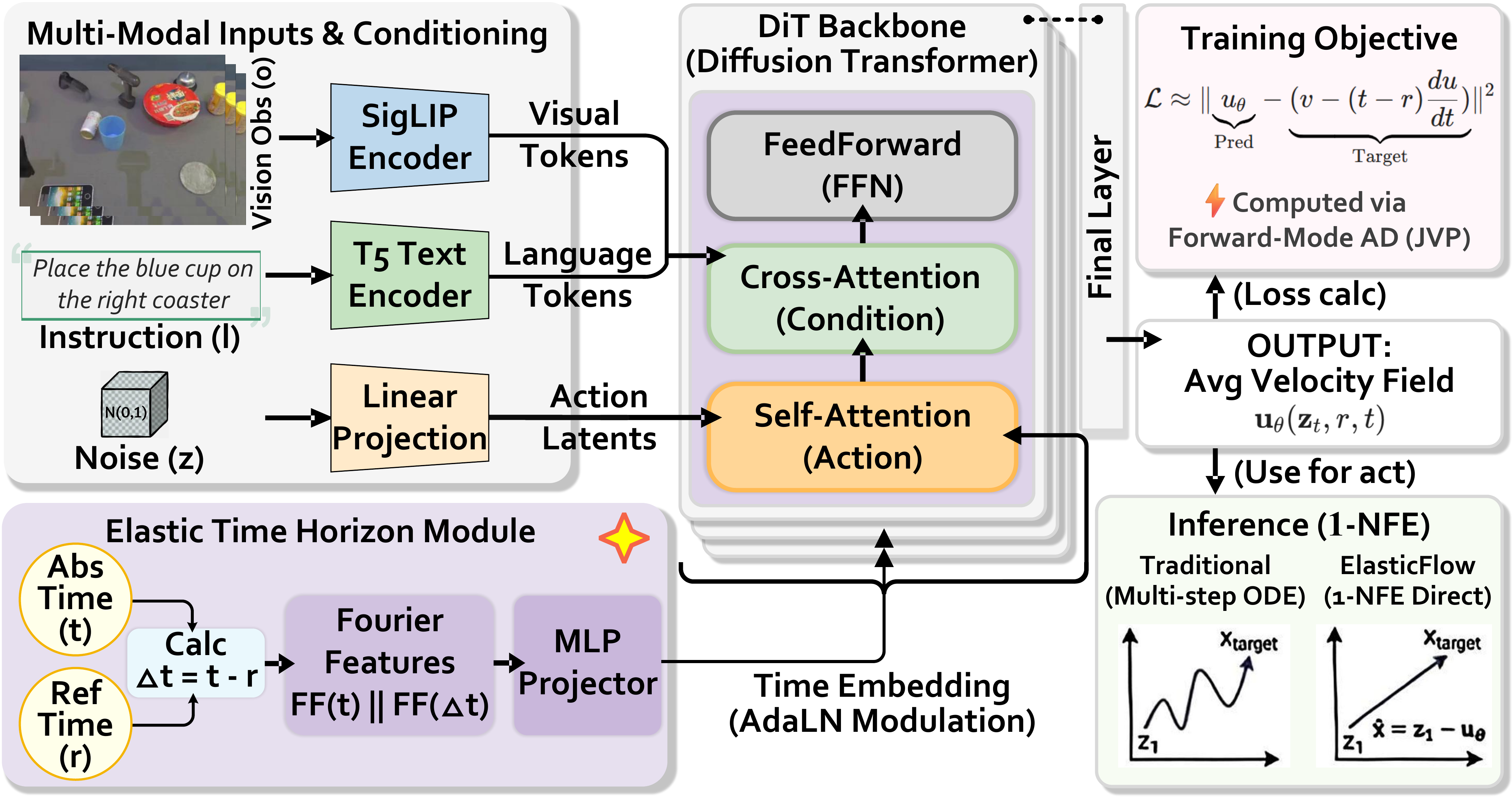}
	\caption{\textbf{ElasticFlow Architecture Overview.} \textbf{Left:} Multi-modal inputs are processed via SigLIP and T5 encoders. The core \textbf{Elastic Time Horizon Module} encodes the time span $\Delta t = t-r$ into Fourier features and injects them into the DiT backbone via AdaLN modulation, thereby explicitly regulating the generated control granularity.
		\textbf{Middle:} The DiT-based backbone network fuses visual and language conditions through cross-attention mechanisms.
		\textbf{Right:} The model predicts the average velocity field $u_\theta$ and uses the \textbf{MeanFlow Identity Loss} based on Forward-Mode AD for supervision. The bottom right shows that this architecture achieves physically consistent \textbf{1-NFE one-step inference}, eliminating the latency of traditional multi-step iterations.}
	\label{fig:architecture}
	\vspace{-8pt}
\end{figure*}

To handle multi-level semantic instructions in robotic tasks (e.g., varying time spans from fine pose adjustments to long-range planning), we design a policy network $u_\theta(z_t, r, t, o, \ell)$ supporting \textbf{Elastic Time Abstraction}. As shown in Figure \ref{fig:architecture}, the overall architecture of ElasticFlow aims to tightly fuse the robot's multi-modal perception with elastic time control.

\paragraph{Dual-Parameter Time Encoding} The network receives not only the absolute flow time $t$ but also the time span $\Delta t = t-r$. We employ Fourier Feature Mapping to encode and fuse both to mitigate the Spectral Bias problem in neural networks when learning high-frequency time variations:

\begin{equation}	
	\text{Emb}(r, t) = \text{MLP}([\text{FF}(t), \text{FF}(t-r)])
\end{equation}

Here, $\text{FF}(\cdot)$ denotes Gaussian Fourier Feature Encoding, $[\cdot, \cdot]$ denotes vector concatenation, and $\text{MLP}$ is a Multi-Layer Perceptron.

Introducing $\Delta t$ aims to mitigate Spectral Bias during learning. More importantly, this dual-parameter encoding endows the model with explicit perception of the "Control Horizon": when $\Delta t$ is small, the model focuses on local high-frequency pose adjustments; when $\Delta t$ is large, the model focuses on long-range trajectory planning. This allows a single model to elastically adapt to control requirements at different levels under unified weights.
Furthermore, this elastic time horizon mechanism enables the model to adaptively determine the required temporal horizon $T$ for a given task. During inference, the continuous flow generated by the policy is discretized into $N$ executable control steps based on the target control frequency, yielding an effective physical step size of $\delta t = T/N$. This formulation allows ElasticFlow to dynamically adjust its execution granularity, effectively bridging high-level semantic complexity with low-level physical control horizons.

\paragraph{Vision-Language Condition Injection} Visual observations $o$ are extracted via the SigLIP encoder, and language instructions $\ell$ are embedded via the text encoder. Feature fusion adopts a DiT (Diffusion Transformer) based backbone architecture, injecting $o$ and $\ell$ into the action flow generation process through Cross-Attention layers.

\begin{figure*}[t]
	\centering
	\footnotesize
	\includegraphics[width=\linewidth]{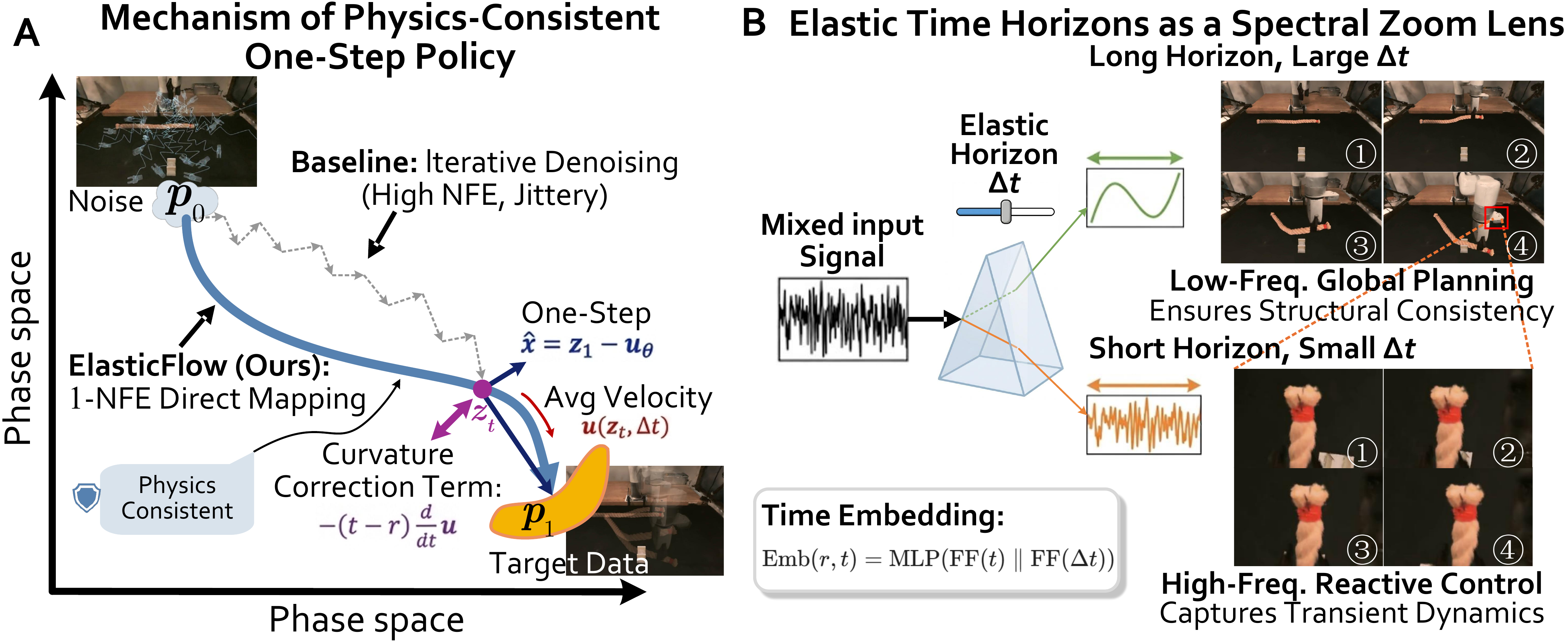}
	\caption{\textbf{Schematic of ElasticFlow Core Mechanisms.} \textbf{(A) Physically Consistent One-Step Geometry:} Unlike iterative denoising (gray), ElasticFlow learns an average velocity field $u$ (blue). This field integrates instantaneous velocity with a \textit{curvature correction term} (purple), naturally ensuring physical consistency and smoothness in one-step generation.
		\textbf{(B) Elastic Time Horizon as a Spectral Zoom Lens:} Addressing Spectral Bias, $\Delta t$ acts as a zoom lens: a smaller $\Delta t$ (top) captures high-frequency reactive control (orange), while a larger $\Delta t$ (bottom) acts as a low-pass filter (green) for long-range planning. This enables a single policy to simultaneously capture transient dynamics and long-horizon structures.}
	\label{fig:mechanism}
	\vspace{-8pt}
\end{figure*}

\subsection{Classifier-Free Guided One-Step Inference and Training}

\label{subsec:training_inference}

\paragraph{Training Objective} Our training objective is to minimize the difference between the network prediction and the target derived from the ElasticFlow Identity. We adopt a \textbf{Classifier-Free Guidance (CFG)} strategy, setting the condition $\ell$ to null (i.e., $c \in \{\ell, \emptyset\}$) with probability $p_{drop}$ during training, thereby jointly learning unconditional and conditional flows. Let $v(z_t, t)$ be the marginal instantaneous velocity field. The loss function is defined as:

\begin{equation}
	\mathcal{L}(\theta) = \mathbb{E}_{t, r, x_1, \epsilon, c}
	\left[ \left\| u_\theta(z_t, r, t, o, c) - \text{sg}\left( \mathcal{T}_{\text{target}} \right) \right\|_2^2 \right]
\end{equation}

Here, $x_1 \sim p_{\text{data}}$ is the target data distribution, $\epsilon \sim \mathcal{N}(0, I)$ is standard Gaussian noise, and $z_t$ is the interpolated state on the flow. $\text{sg}(\cdot)$ denotes the Stop-Gradient operator, used to stabilize the bootstrapped training objective. The target term $\mathcal{T}_{\text{target}}$ is defined as:

\begin{equation}
	\mathcal{T}_{\text{target}} = v(z_t, t) - (t-r)\left( v(z_t, t) \cdot \nabla_z u_\theta + \partial_t u_\theta \right)
\end{equation}

In this equation, $\nabla_z u_\theta$ is the Jacobian matrix of $u_\theta$ with respect to state $z$, and $\partial_t u_\theta$ is the partial derivative with respect to time $t$. We utilize Forward-mode Automatic Differentiation (Forward-mode AD) to efficiently compute the Jacobian-Vector Product (JVP) on the right side, avoiding the high computational overhead of the Hessian matrix.

\paragraph{Efficient One-Step Inference (1-NFE Inference)}

In the inference phase, thanks to the properties of the average velocity field, we do not need to perform dozens of iterations like traditional diffusion policies. Given initial Gaussian noise $z_1 \sim \mathcal{N}(0, I)$, we directly set $r=0, t=1$. Through a single network forward pass, we can obtain the denoised action trajectory $\hat{x}$:

\begin{equation}
	\hat{x} = z_1 - \left( u_\theta(\cdot, \emptyset) + w \cdot (u_\theta(\cdot, \ell) - u_\theta(\cdot, \emptyset)) \right)
\end{equation}

Here, $z_1 \sim \mathcal{N}(0, I)$ is the initial noise, and $w$ is the Guidance Scale for Classifier-Free Guidance (CFG), used to balance the semantic alignment and diversity of generation. This feature boosts the policy inference frequency to over 100Hz, ensuring real-time responsiveness to dynamic environments in complex semantic scenarios.

\section{Experiments}
\label{sec:experiments}

To comprehensively evaluate the effectiveness of the proposed ElasticFlow framework, we conducted extensive experiments on three challenging robot manipulation benchmarks: LIBERO~\citep{liu2023libero}, CALVIN~\citep{mees2022calvin}, and RoboTwin~\citep{mu2025robotwin}. These benchmarks cover a wide range of manipulation skills, task horizon lengths, and semantic complexity. Notably, the inclusion of high-fidelity simulators aligns with recent trends in bridging the real-to-sim gap for reliable evaluation, such as utilizing Gaussian Splatting~\citep{zhang2025real}. Additionally, to verify the model's robustness in the physical world, we deployed ElasticFlow for real-robot manipulation evaluation. Detailed training and inference hardware configurations are provided in Appendix~\ref{app:hardware}.

Our experiments aim to answer three core questions: \textbf{(1) Generalization}: Can ElasticFlow achieve efficient generalization across diverse tasks and spatial configurations? \textbf{(2) Long-Horizon Consistency}: Can the proposed Elastic Time Horizon mechanism effectively mitigate error accumulation in long-range tasks? \textbf{(3) Continuous Instruction Following}: Can the model maintain performance stability amidst continuously changing natural language instruction streams?

\subsection{LIBERO Benchmark Performance}

We first evaluate our method on the LIBERO benchmark, widely used for assessing the lifelong learning and generalization capabilities of robot policies. LIBERO consists of four suites: \textit{LIBERO-Spatial} (spatial generalization), \textit{LIBERO-Object} (object generalization), \textit{LIBERO-Goal} (goal-conditioned generalization), and the highly challenging \textit{LIBERO-Long}, which involves long-horizon sequential task chains.

\paragraph{Overall Performance}
Table~\ref{tab:libero_all} summarizes the success rates for all four task suites. ElasticFlow achieved a new state-of-the-art (SOTA) average success rate of \textbf{98.5\%}, surpassing the previous best method HiF-VLA (98.0\%) and significantly outperforming diffusion-based baselines such as $\pi_0$ (94.2\%) and Octo (75.1\%). Notably, the high success rates demonstrated by our method in \textit{LIBERO-Object} (99.3\%) and \textit{LIBERO-Goal} (98.7\%) prove ElasticFlow's high precision in object manipulation and goal-conditioned planning.

\begin{table*}[ht]
	\footnotesize
	\centering
	\renewcommand{\arraystretch}{0.95}
	\caption{Overall performance on \textbf{LIBERO} benchmark suites. We compare our method against a range of state-of-the-art algorithms. Best performance is highlighted in \textbf{bold}.}
	\label{tab:libero_all}
	\resizebox{\linewidth}{!}{
		\begin{tabular}{cccccc}
			\toprule
			\textbf{Methods} &\textbf{LIBERO-Spatial}& \textbf{LIBERO-Object }&\textbf{LIBERO-Goal}& \textbf{LIBERO-Long}& \textbf{Average}\\
			\midrule
			TraceVLA~\cite{zhengtracevla}&84.6&85.2& 75.1 & 54.1 & 74.8\\
			Octo~\cite{team2024octo} & 78.9 & 85.7 & 84.6 & 51.1 & 75.1\\
			CoT-VLA~\cite{zhao2025cot} & 81.1 & 87.5 & 91.6 & 87.6 & 69.0\\
			SpatialVLA~\cite{li2025spatial} &88.2&89.9&78.6&55.5&78.1\\
			ThinkAct~\cite{huangthinkact}& 88.3 & 91.4 & 87.1 & 70.9 &84.4 \\
			Seer~\cite{tianpredictive} &-&-&-&87.7&87.7\\
			FlowVLA~\cite{zhong2025flowvla} & 93.2 & 95.0 & 91.6 & 72.6 & 88.1\\
			DreamVLA~\cite{zhangdreamvla} &97.5& 94.0 & 89.5 & 89.5 &92.6 \\
			CogACT~\cite{li2024cogact} & 97.2 & 98.0 & 90.2 & 88.8 & 93.2\\ 
			$\pi_0$~\cite{black2024pi_0} & 96.8 & 98.8 & 95.8 & 85.2 &94.2 \\
			GR00T N1~\cite{bjorck2025gr00t} &94.4&97.6&93.0&90.6&93.9\\
			UniVLA~\cite{bu2025univla} & 96.5 & 96.8 & 95.6 & 92.0 &95.2 \\
			MemoryVLA~\cite{shi2025memoryvla}& 98.4 & 98.4 & 96.4 &93.4 & 96.5\\
			OpenVLA-OFT~\cite{kim2025fine} & 97.6 & 98.4 & 97.9 & 94.5 &97.1 \\
			HiF-VLA~\citep{lin2025hifvla} & \textbf{98.8} & \textbf{99.4}& 97.4 & 96.4 & 98.0 \\
			\rowcolor{gray!10}
			\textbf{ElasticFlow (Ours)} & 98.4 & 99.3 & \textbf{98.7} & \textbf{97.6} & \textbf{98.5} \\
			\bottomrule
		\end{tabular}
	}
	\vspace{-8pt}
\end{table*}

\paragraph{Long-Horizon Task Analysis}
To further investigate the impact of our average velocity field modeling on long-horizon tasks, we present detailed breakdown results for the \textit{LIBERO-Long} suite in Table~\ref{tab:libero_long} in Appendix. 

\subsection{Continuous Instruction Following on CALVIN}

The CALVIN benchmark evaluates an agent's ability to follow streams of natural language instructions in a continuous environment (ABC-D split). The key metric is the average number of successfully completed instructions in a row (Avg. Chain Length).

Table~\ref{tab:third_multi_view_comparison} reports performance under third-person and multi-view settings. ElasticFlow sets a new benchmark, achieving an average chain length of \textbf{4.37} in multi-view settings and \textbf{4.15} in single-view settings.
Unlike methods whose performance degrades after the first few tasks, ElasticFlow maintains extremely high success rates even at the 4th and 5th instructions (83.6\% and 72.7\%, respectively). This indicates that the global consistency enforced by the MeanFlow Identity enables the policy to transition smoothly between different semantic goals without losing physical control stability, effectively handling state drift caused by long sequences of instructions.

\begin{table*}[t]
	\footnotesize
	\centering
	\setlength{\tabcolsep}{5pt}
	\renewcommand{\arraystretch}{0.95}
	\caption{Performance comparison on the \textbf{CALVIN ABC-D} benchmark. We report the average number of successfully completed tasks in a row (Chain Length). \textbf{Bold} indicates the best performance.}
	\label{tab:third_multi_view_comparison}
	\begin{tabular}{l|lccccc|c}
		\toprule
		\textbf{View} & \textbf{Method} & 1 & 2 & 3 & 4 & 5 & \textbf{Avg. Len.} $\uparrow$ \\
		\midrule
		\multirow{7}{*}{Third View}
		& SuSIE \cite{blackzero} & 87.0 & 69.0 & 49.0 & 38.0 & 26.0 & 2.69 \\
		& OpenVLA \cite{kim2025openvla} & 91.3 & 77.8 & 62.0 & 52.1 & 43.5 & 3.27 \\
		& CLOVER \cite{bu2024closed} & \textbf{{96.0}} & 83.5 & 70.8 & 57.5 & 45.4 & 3.53 \\
		& VPP \cite{huvideo} &90.9&81.5&71.3&62.0&51.8&3.58\\
		& $\pi_0$ \cite{black2024pi_0} &93.7&83.2&74.0&62.9&51.0&3.65\\
		& UniVLA \cite{bu2025univla} & 95.5 & 85.8 & 74.8 & 66.9 & 56.5 & 3.80 \\
		& HiF-VLA\cite{lin2025hifvla} & 93.5 & 87.4 & \textbf{81.4} & 75.9 & 69.4 & 4.08 \\
		\rowcolor{gray!10}
		& \textbf{ElasticFlow (Ours)} & 95.6 & \textbf{88.7} & 79.8 & \textbf{77.3} & \textbf{73.4} & \textbf{4.15} \\
		\midrule
		\multirow{9}{*}{Multi-View}
		& GR-1 \cite{wuunleashing} & 85.4 & 71.2 & 59.6 & 49.7 & 40.1 & 3.06 \\
		& Vidman \cite{wen2024vidman} & 91.5 & 76.4 & 68.2 & 59.2 & 46.7 & 3.42 \\
		& $\pi_0$ \cite{black2024pi_0} & 93.8 & 85.0 & 76.7 & 68.1 & 59.9 & 3.92 \\
		& UP-VLA \cite{zhang2025up} & 92.8 & 86.5 & 81.5 & 76.9 & 69.9 & 4.08 \\
		& OpenVLA-OFT \cite{kim2025fine} & 96.3 & 89.1 & 82.4 & 75.8 & 66.5 & 4.10 \\
		& RoboVLMs \cite{liu2024towards} &98.0&93.6 &85.4&77.8&70.4&4.25 \\
		& Seer \cite{tianpredictive} & 96.3 & 91.6 & 86.1 & 80.3 & 74.0 & 4.28 \\
		& VPP \cite{huvideo} & 96.5 & 90.9 & 86.6 & 82.0 & \textbf{76.9} & 4.33 \\
		& HiF-VLA~\citep{lin2025hifvla} & \textbf{98.5} & 94.1 & \textbf{88.1} & 81.4 & 73.1 & 4.35 \\
		\rowcolor{gray!10}
		& \textbf{ElasticFlow (Ours)} & 97.8 & \textbf{95.3} & 87.9 & \textbf{83.6} & 72.7 & \textbf{4.37} \\
		\bottomrule
	\end{tabular}
	\vspace{-8pt}
\end{table*}

\subsection{Horizon Dependency Analysis on RoboTwin}

To explicitly verify the "elastic" nature of our framework, we conducted a group analysis on RoboTwin2.0 based on task horizon length: Short (100-130 steps), Medium (150-230 steps), and Long/Extra Long (280-650 steps). Figure~\ref{fig:robotwin_viz} in Appendix visualizes the execution process.
As detailed in Appendix Table~\ref{tab:robotwin2_results_by_difficulty}, results reveal that while baselines like SimpleVLA-RL decay on long horizons, ElasticFlow demonstrates significant stability. Specifically, in the \textit{Long \& Extra Long} category, ElasticFlow achieved \textbf{71.1\%}, outperforming SimpleVLA-RL (69.0\%), $\pi_0$ (43.3\%), and RDT (27.8\%). This confirms that explicitly encoding $\Delta t$ enables the network to adaptively adjust its generation strategy—balancing reactive control in short tasks with structural coherence in long-duration operations.

To further verify generalization in high-fidelity simulation, we conducted a qualitative analysis on \textbf{RoboCasa}. As shown in Appendix Figure~\ref{fig:robocasa_viz}, even in kitchen tasks involving complex contact dynamics, ElasticFlow generates smooth, stable, and temporally consistent action sequences.

\subsection{Real-World Experiments}
\label{sec:real_world}

\begin{figure*}[t]
	\centering
	\includegraphics[width=\linewidth]{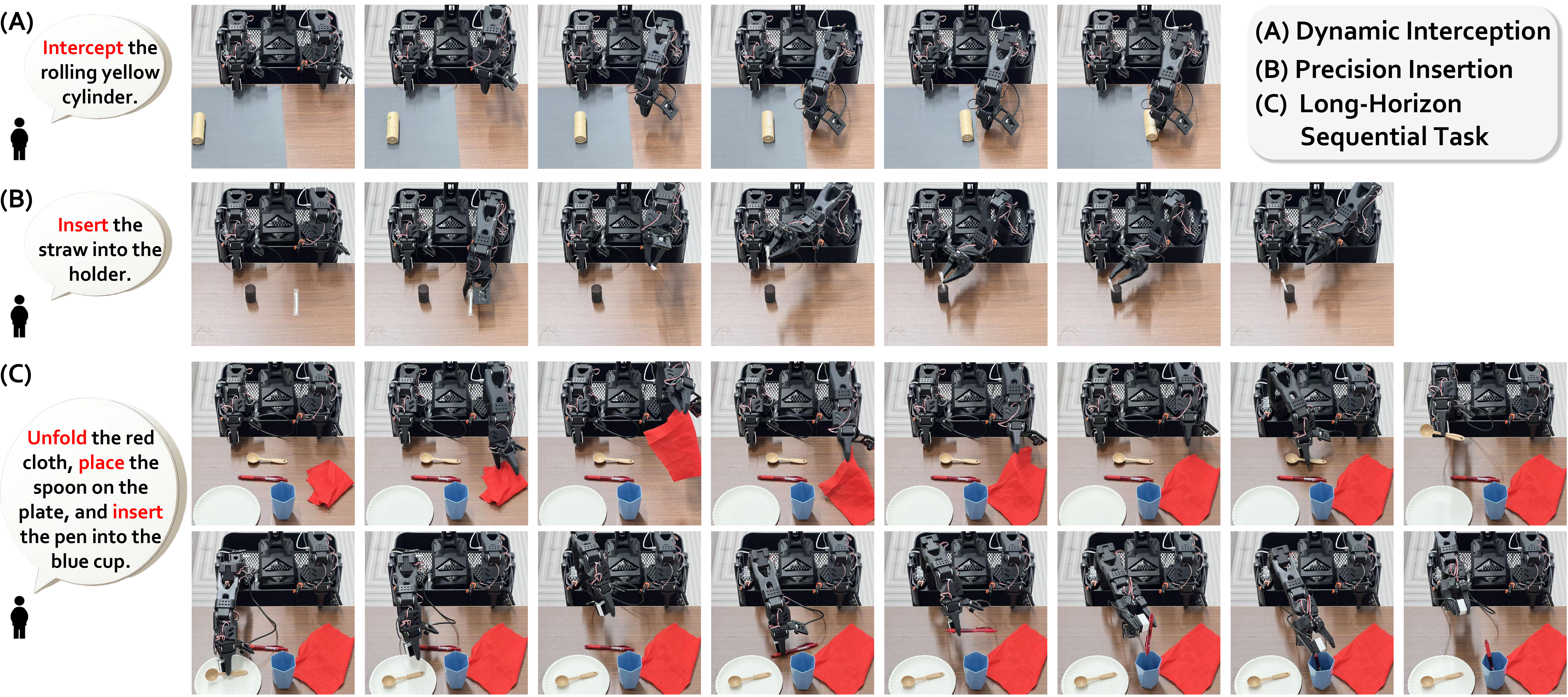}
	\caption{\textbf{Qualitative Evaluation of ElasticFlow on Real Robots.} We tested the model's performance in the real world on XLeRobot.
		\textbf{(A) Dynamic Interception:} Intercepting a fast-rolling cylinder verifies the 71Hz response. 
		\textbf{(B) Precision Assembly:} Deformable straw insertion demonstrates physical smoothness. 
		\textbf{(C) Long-Horizon Sequential Manipulation:} Elastic horizons ensure structural consistency in multi-stage tasks.}
	\label{fig:real_world_experiments}
	\vspace{-5pt}
\end{figure*}

To validate ElasticFlow's capability to cross the Sim-to-Real gap, we deployed the trained policy on a real robotic arm platform (UFACTORY xArm6) for qualitative and quantitative evaluation. We designed seven tasks covering various control challenges (e.g., high-frequency dynamic response, precision assembly, long-horizon multi-stage manipulation). Figure~\ref{fig:real_world_experiments} illustrates the execution process of three representative tasks. For detailed definitions, evaluation protocols, and quantitative statistics of all 7 real-world tasks, please refer to Appendix~\ref{app:real_world_details} and Table~\ref{tab:real_world_comprehensive}.

\subsection{Ablation Studies and Mechanism Analysis}
\label{subsec:ablation}

To deeply analyze the contribution of each core component in the ElasticFlow framework, we conducted a series of controlled ablation experiments on the "Long Horizon" task subset of the RoboTwin benchmark. We focused on verifying the following three hypotheses: (1) Elastic Time Horizon is necessary for handling multi-stage tasks; (2) The training objective derived from MeanFlow Identity is superior to the standard Flow Matching objective; (3) Classifier-Free Guidance (CFG) remains effective in single-step inference.

\vspace{-5pt}
\paragraph{Effectiveness of Elastic Time Horizon}
To verify the necessity of the elastic time horizon mechanism, we compared the full ElasticFlow model with two variants. Specifically, the \textbf{w/o Horizon Input} variant removes the $\Delta t$ time embedding from the network input, retaining only absolute time $t$, to assess the contribution of explicit time span information. The \textbf{Fixed Horizon} variant sets $\Delta t$ to a constant (e.g., corresponding to a 10-step action chunk) during both training and inference, no longer dynamically adjusting with task stages, to investigate the impact of horizon adaptability.

The experimental results are shown in Table~\ref{tab:ablation_horizon}. Results indicate that removing horizon information leads to an 18.4\% drop in success rate for long-range tasks, proving that explicit perception of "control granularity" is crucial for long-sequence action planning.
\textbf{To further verify the "Spectral Zoom Lens" metaphor in Figure~\ref{fig:mechanism}(B), we conducted a Mismatch Test.} As shown, forcing a small $\Delta t$ input during long-horizon tasks (i.e., forcing the model to focus only on the immediate) caused myopic behavior, failing to complete global planning (success rate only 45.3\%). Conversely, forcing a large $\Delta t$ input during short-horizon reactive tasks led to sluggish response (success rate dropped to 55.7\%). These results strongly confirm the physical necessity of the elastic horizon mechanism.

\begin{table}[h]
	\centering
	\caption{\textbf{Ablation and Mismatch Analysis of Elastic Time Horizon Mechanism.} Besides comparing with Fixed Horizon, we conducted a \textbf{Mismatch Test}: forcing an incorrect $\Delta t$ during inference. Results confirm that short-horizon input cannot guide long-range planning, while long-horizon input causes sluggish short-range reaction, validating the physical meaning of the "Spectral Zoom Lens" metaphor.}
	\label{tab:ablation_horizon}
	\setlength{\tabcolsep}{1pt} 
	\renewcommand{\arraystretch}{1.1}
	\resizebox{\linewidth}{!}{
		\begin{tabular}{llccc}
			\toprule
			\textbf{Model Variant} & \textbf{Setting} & \textbf{Long Horizon SR} & \textbf{Short Horizon SR} & \textbf{Note} \\
			\midrule
			w/o Horizon Input & N/A & 52.7\% & 61.5\% & - \\
			Fixed Horizon & $\Delta t=10$ (Fixed) & 58.2\% & 94.5\% & Fails Long \\
			Fixed Horizon & $\Delta t=50$ (Fixed) & 62.1\% & 55.4\% & Fails Short \\
			\midrule
			\textit{Mismatch Test} & \textit{Force $\Delta t=10$ on Long} & 45.3\% & - & \textit{Myopic} \\
			\textit{Mismatch Test} & \textit{Force $\Delta t=50$ on Short} & - & 55.7\% & \textit{Sluggish} \\
			\midrule
			\rowcolor{gray!10}
			\textbf{ElasticFlow (Full)} & \textbf{Dynamic $\Delta t$} & \textbf{71.1\%} & \textbf{98.2\%} & \textbf{Optimal} \\
			\bottomrule
		\end{tabular}
	}
	\vspace{-10pt}
\end{table}

\vspace{-5pt}
\paragraph{Comparison of Training Objectives: MeanFlow vs. Flow Matching}
To verify the necessity of the MeanFlow Identity in Equation (\ref{eq:ElasticFlow_identity}), we trained a baseline model using the same network architecture but with the standard Conditional Flow Matching (CFM) loss function, which directly regresses the instantaneous velocity field $v_t$ instead of the average velocity field $u_t$.

As shown in Table~\ref{tab:ablation_objective}, under the 1-NFE inference setting, the standard CFM model fails to generate effective actions (success rate only 12.4\%). Additionally, to quantitatively evaluate physical consistency, we introduced the Trajectory Jerk metric, defined as $\mathcal{J} = \frac{1}{T} \sum_{t} \| \dddot{x}_t \|^2$, where $T$ is the total number of time steps in the trajectory and $\dddot{x}_t$ represents the third derivative of the end-effector position at time $t$. Results show that even with 10-step iteration, CFM's trajectory smoothness ($3.2 \times 10^{-3}$) is inferior to ElasticFlow's single-step result ($1.1 \times 10^{-3}$). This proves that the curvature correction term implicit in the MeanFlow Identity effectively suppresses high-frequency jitter, achieving better physical compliance.

\begin{table}[h]
	\centering
	\caption{\textbf{Impact of Training Objective, Smoothness, and Inference Steps.} Besides success rate, we report \textbf{Trajectory Smoothness (Avg. Jerk $\downarrow$)}. This metric is calculated as the Mean Squared Third Derivative of the end-effector position with respect to time. Results indicate that ElasticFlow generates smoother trajectories under 1-NFE than standard CFM, verifying its physical consistency.}
	\label{tab:ablation_objective}
	\setlength{\tabcolsep}{2.5pt}
	\renewcommand{\arraystretch}{1.1}
	\resizebox{\linewidth}{!}{
		\begin{tabular}{lccccc}
			\toprule
			\textbf{Training Objective} & \textbf{Steps} & \textbf{Success Rate} & \textbf{Smoothness (Jerk $\downarrow$)} & \textbf{Latency} \\
			\midrule
			Standard CFM ($v_t$) & 1-NFE & 12.4\% & $8.5 \times 10^{-2}$ & 14ms \\
			Standard CFM ($v_t$) & 10-NFE & 68.5\% & $3.2 \times 10^{-3}$ & 140ms \\
			\rowcolor{gray!10}
			\textbf{ElasticFlow ($u_t$)} & \textbf{1-NFE} & \textbf{71.1\%} & \textbf{1.1 $\times$ 10$^{-3}$} & \textbf{14ms} \\
			\bottomrule
		\end{tabular}
	}
	\vspace{-10pt}
\end{table}

Finally, in \textbf{Appendix~\ref{app:cfg_ablation}}, we further analyze the impact of Classifier-Free Guidance (CFG) weight $w$ on model performance. Results indicate that ElasticFlow maintains robust performance across a wide range of weights ($w \in [1.5, 2.5]$).

\subsection{Inference Latency and Real-Time Analysis}
\label{subsec:latency}

In embodied AI applications, policy Inference Latency directly determines the control loop frequency, thereby affecting the robot's responsiveness to dynamic environments. We measured the inference speed of different policy models on an NVIDIA RTX 4090 GPU. The test input was a $224 \times 224$ image observation and a natural language instruction, with a Batch Size of 1.

Table~\ref{tab:latency_comparison} reports the Inference Time per Step and Max Control Hz for different methods. Auto-regressive models like OpenVLA are limited by parameter size (7B) and token-by-token generation, maintaining only a low-frequency inference of about 5Hz, which is difficult to meet high-frequency control needs. Traditional Diffusion Policy is also constrained by the computational overhead of multi-step iterative denoising (DDIM), with latency generally exceeding 100ms. ElasticFlow breaks this efficiency bottleneck with its 1-NFE single-step generation mechanism, drastically reducing inference latency to 14ms (\textbf{71Hz}). This metric represents a 5x and 14x improvement over Diffusion Policy and OpenVLA, respectively, enabling easy deployment in dynamic manipulation tasks requiring high real-time performance.

\begin{table}[h]
	\centering
	\caption{\textbf{Comparison of Inference Latency and Control Frequency.} All tests were conducted on the same NVIDIA RTX 4090. Max Control Frequency (Hz) is calculated as $1000 / \text{Latency (ms)}$. ElasticFlow achieves the lowest latency while ensuring high success rates, sitting on the speed-accuracy Pareto Frontier.}
	\label{tab:latency_comparison}
	\setlength{\tabcolsep}{1pt}
	\renewcommand{\arraystretch}{1.2}
	\resizebox{\linewidth}{!}{
		\begin{tabular}{lcccc}
			\toprule
			\textbf{Method} & \textbf{Backbone} & \textbf{NFE} & \textbf{Time (ms)} $\downarrow$ & \textbf{Freq (Hz)} $\uparrow$ \\
			\midrule
			OpenVLA~\cite{kim2025openvla} & Transformer (7B) & Auto-reg. & 200.0 & $\sim$ 5 \\
			Diffusion Policy~\cite{chi2023diffusion} & UNet (300M) & 16 (DDIM) & 120.0 & $\sim$ 8 \\
			$\pi_0$~\cite{black2024pi_0} & DiT (300M) & 10 (Euler) & 85.0 & $\sim$ 12 \\
			Consistency Policy~\cite{prasad2024consistency} & UNet (300M) & 2 & 28.0 & $\sim$ 35 \\
			\rowcolor{gray!10}
			\textbf{ElasticFlow (Ours)} & DiT (150M) & \textbf{1} & \textbf{14.0} & \textbf{$\sim$ 71} \\
			\bottomrule
		\end{tabular}
	}
	\vspace{-8pt}
\end{table}

\section{Conclusion}
\label{sec:conclusion}

We proposed \textbf{ElasticFlow}, a physics-consistent one-step policy framework addressing the latency and inconsistency of diffusion policies. By modeling the average velocity field via Mean Field Theory, we enable direct, distillation-free generation from noise to action. Additionally, our \textbf{Elastic Time Horizons} mechanism explicitly encodes control granularity to overcome spectral bias, bridging high-frequency reflexes and long-range planning. Empirically, ElasticFlow achieves $\sim$71Hz 1-NFE inference and outperforms state-of-the-art methods like OpenVLA on benchmarks including LIBERO-Long and CALVIN. This work offers a robust, efficient paradigm for real-time embodied generative modeling.

\section{Limitations}
\label{sec:limitations}

Although ElasticFlow performed exceptionally in experiments, this study still has limitations, which point to directions for future work:

\textbf{(1) Scaling with Large-Scale Pre-training}: Currently, our experiments mainly focus on data scales of millions of interactions. In the future, we will explore the Scaling Laws of ElasticFlow on billion-scale cross-embodiment datasets (e.g., Open X-Embodiment) to verify its potential as a core generation head for general-purpose Robot Foundation Models.

\textbf{(2) Closed-Loop Control and Online Correction}: Thanks to its extremely high inference frequency, ElasticFlow is suitable for high-frequency closed-loop control. Future work can combine Model Predictive Control (MPC) or online Reinforcement Learning, using its rapidly generated trajectories as Warm-start initial values to further improve the robot's robustness in unstructured environments.

\textbf{(3) Deepening Fine-Grained Semantic Alignment}: Current language guidance is mainly injected via cross-attention. Future exploration could involve deeper latent space fusion of ElasticFlow with Multimodal Large Language Models (MLLMs), \textbf{leveraging ElasticFlow's time elasticity to better parse and execute long-range, multi-stage complex natural language instructions.}

\section{Ethical Considerations}
\label{sec:ethics}

This study was conducted in standard simulation environments and real-world physical environments, involving no sensitive personal data. Regarding potential real-world deployment, we focus on the following ethical dimensions:

\paragraph{Physical Safety and Green Computing}
High-frequency jitter in traditional diffusion policies may pose physical safety hazards. ElasticFlow effectively suppresses non-physical oscillations through theoretical curvature correction, enhancing the safety of human-robot collaboration. Meanwhile, its 1-NFE single-step inference mechanism significantly reduces computational energy consumption throughout the robot's lifecycle, adhering to the sustainable development principles of Green AI. Nevertheless, hardware emergency stop mechanisms must still be equipped during Sim-to-Real transfer to cope with dynamic discrepancies.

\paragraph{Social Responsibility and Misuse}
As the system integrates pre-trained T5 models, it may inherit potential data biases. When extending to the open world, strict semantic alignment and instruction filtering mechanisms (Safety Guardrails) must be established to prevent the model from executing harmful instructions or being used for unethical purposes (e.g., weaponization), ensuring technology serves only benign goals aiding human production and life.

\section*{Acknowledgments}

This work was supported by the National Natural Science Foundation of China under Grant 62372427, in part by Chongqing Natural Science Foundation Innovation and Development Joint Fund (No. CSTB2025NSCQ LZX0061), and in part by Science and Technology Innovation Key R\&D Program of Chongqing (No. CSTB2025TIAD-STX0023).

\bibliography{strings,custom}

\begin{thebibliography}{40}
\providecommand{\natexlab}[1]{#1}

\bibitem[{Bjorck et~al.(2025)Bjorck, Casta{\~n}eda, Cherniadev, Da, Ding, Fan,
  Fang, Fox, Hu, Huang et~al.}]{bjorck2025gr00t}
Johan Bjorck, Fernando Casta{\~n}eda, Nikita Cherniadev, Xingye Da, Runyu Ding,
  Linxi Fan, Yu~Fang, Dieter Fox, Fengyuan Hu, Spencer Huang, and 1 others.
  2025.
\newblock Gr00t n1: An open foundation model for generalist humanoid robots.
\newblock \emph{arXiv preprint arXiv:2503.14734}.

\bibitem[{Black et~al.(2024{\natexlab{a}})Black, Brown, Driess, Esmail, Equi,
  Finn, Fusai, Groom, Hausman, Ichter et~al.}]{black2024pi_0}
Kevin Black, Noah Brown, Danny Driess, Adnan Esmail, Michael Equi, Chelsea
  Finn, Niccolo Fusai, Lachy Groom, Karol Hausman, Brian Ichter, and 1 others.
  2024{\natexlab{a}}.
\newblock $pi\_0 $: A vision-language-action flow model for general robot
  control.
\newblock \emph{arXiv preprint arXiv:2410.24164}.

\bibitem[{Black et~al.(2024{\natexlab{b}})Black, Nakamoto, Atreya, Walke, Finn,
  Kumar, and Levine}]{blackzero}
Kevin Black, Mitsuhiko Nakamoto, Pranav Atreya, Homer~Rich Walke, Chelsea Finn,
  Aviral Kumar, and Sergey Levine. 2024{\natexlab{b}}.
\newblock Zero-shot robotic manipulation with pre-trained image-editing
  diffusion models.
\newblock In \emph{The Twelfth International Conference on Learning
  Representations}.

\bibitem[{Bu et~al.(2025)Bu, Yang, Cai, Gao, Ren, Yao, Luo, and
  Li}]{bu2025univla}
Qingwen Bu, Yanting Yang, Jisong Cai, Shenyuan Gao, Guanghui Ren, Maoqing Yao,
  Ping Luo, and Hongyang Li. 2025.
\newblock Univla: Learning to act anywhere with task-centric latent actions.
\newblock \emph{arXiv preprint arXiv:2505.06111}.

\bibitem[{Bu et~al.(2024)Bu, Zeng, Chen, Yang, Zhou, Yan, Luo, Cui, Ma, and
  Li}]{bu2024closed}
Qingwen Bu, Jia Zeng, Li~Chen, Yanchao Yang, Guyue Zhou, Junchi Yan, Ping Luo,
  Heming Cui, Yi~Ma, and Hongyang Li. 2024.
\newblock Closed-loop visuomotor control with generative expectation for
  robotic manipulation.
\newblock \emph{Advances in Neural Information Processing Systems},
  37:139002--139029.

\bibitem[{Chi et~al.(2023)Chi, Feng, Du, Xu, Cousineau, Burchfiel, and
  Song}]{chi2023diffusion}
Cheng Chi, Siyuan Feng, Yilun Du, Zhenjia Xu, Eric Cousineau, Benjamin
  Burchfiel, and Shuran Song. 2023.
\newblock Diffusion policy: Visuomotor policy learning via action diffusion.
\newblock In \emph{Robotics: Science and Systems}.

\bibitem[{Geng et~al.(2025)Geng, Deng, Bai, Kolter, and He}]{geng2025mean}
Zhengyang Geng, Mingyang Deng, Xingjian Bai, J~Zico Kolter, and Kaiming He.
  2025.
\newblock Mean flows for one-step generative modeling.
\newblock In \emph{The Thirty-ninth Annual Conference on Neural Information
  Processing Systems}.

\bibitem[{Ghosh et~al.(2024)Ghosh, Walke, Pertsch, Black, Mees, Dasari, Hejna,
  Kreiman, Xu, Luo, Tan, Chen, Vuong, Xiao, Sanketi, Sadigh, Finn, and
  Levine}]{team2024octo}
Dibya Ghosh, Homer~Rich Walke, Karl Pertsch, Kevin Black, Oier Mees, Sudeep
  Dasari, Joey Hejna, Tobias Kreiman, Charles Xu, Jianlan Luo, You~Liang Tan,
  Lawrence~Yunliang Chen, Quan Vuong, Ted Xiao, Pannag~R. Sanketi, Dorsa
  Sadigh, Chelsea Finn, and Sergey Levine. 2024.
\newblock Octo: An open-source generalist robot policy.
\newblock In \emph{Robotics: Science and Systems}.

\bibitem[{Hu et~al.(2025)Hu, Guo, Wang, Chen, Wang, Zhang, Sreenath, Lu, and
  Chen}]{huvideo}
Yucheng Hu, Yanjiang Guo, Pengchao Wang, Xiaoyu Chen, Yen-Jen Wang, Jianke
  Zhang, Koushil Sreenath, Chaochao Lu, and Jianyu Chen. 2025.
\newblock Video prediction policy: A generalist robot policy with predictive
  visual representations.
\newblock In \emph{International Conference on Machine Learning}, pages
  24328--24346. PMLR.

\bibitem[{Huang et~al.(2025)Huang, Wu, Chen, Wang, and Yang}]{huangthinkact}
Chi-Pin Huang, Yueh-Hua Wu, Min-Hung Chen, Yu-Chiang~Frank Wang, and Fu-En
  Yang. 2025.
\newblock Thinkact: Vision-language-action reasoning via reinforced visual
  latent planning.
\newblock In \emph{The Thirty-ninth Annual Conference on Neural Information
  Processing Systems}.

\bibitem[{Kim et~al.(2025{\natexlab{a}})Kim, Finn, and Liang}]{kim2025fine}
Moo~Jin Kim, Chelsea Finn, and Percy Liang. 2025{\natexlab{a}}.
\newblock Fine-tuning vision-language-action models: Optimizing speed and
  success.
\newblock \emph{arXiv preprint arXiv:2502.19645}.

\bibitem[{Kim et~al.(2025{\natexlab{b}})Kim, Pertsch, Karamcheti, Xiao,
  Balakrishna, Nair, Rafailov, Foster, Sanketi, Vuong et~al.}]{kim2025openvla}
Moo~Jin Kim, Karl Pertsch, Siddharth Karamcheti, Ted Xiao, Ashwin Balakrishna,
  Suraj Nair, Rafael Rafailov, Ethan~P Foster, Pannag~R Sanketi, Quan Vuong,
  and 1 others. 2025{\natexlab{b}}.
\newblock Openvla: An open-source vision-language-action model.
\newblock In \emph{Conference on Robot Learning}, pages 2679--2713. PMLR.

\bibitem[{Li et~al.(2025{\natexlab{a}})Li, Song, Zhao, Wang, Ding, Wang, Zeng,
  and Li}]{li2025spatial}
Fuhao Li, Wenxuan Song, Han Zhao, Jingbo Wang, Pengxiang Ding, Donglin Wang,
  Long Zeng, and Haoang Li. 2025{\natexlab{a}}.
\newblock Spatial forcing: Implicit spatial representation alignment for
  vision-language-action model.
\newblock \emph{arXiv preprint arXiv:2510.12276}.

\bibitem[{Li et~al.(2025{\natexlab{b}})Li, Zuo, Yu, Zhang, Yang, Zhang, Zhu,
  Zhang, Chen, Cui et~al.}]{li2025simplevla}
Haozhan Li, Yuxin Zuo, Jiale Yu, Yuhao Zhang, Zhaohui Yang, Kaiyan Zhang,
  Xuekai Zhu, Yuchen Zhang, Tianxing Chen, Ganqu Cui, and 1 others.
  2025{\natexlab{b}}.
\newblock Simplevla-rl: Scaling vla training via reinforcement learning.
\newblock \emph{arXiv preprint arXiv:2509.09674}.

\bibitem[{Li et~al.(2024{\natexlab{a}})Li, Liang, Wang, Luo, Chen, Liao, Wei,
  Deng, Xu, Zhang et~al.}]{li2024cogact}
Qixiu Li, Yaobo Liang, Zeyu Wang, Lin Luo, Xi~Chen, Mozheng Liao, Fangyun Wei,
  Yu~Deng, Sicheng Xu, Yizhong Zhang, and 1 others. 2024{\natexlab{a}}.
\newblock Cogact: A foundational vision-language-action model for synergizing
  cognition and action in robotic manipulation.
\newblock \emph{arXiv preprint arXiv:2411.19650}.

\bibitem[{Li et~al.(2025{\natexlab{c}})Li, Gao, Sadigh, and
  Song}]{li2025unified}
Shuang Li, Yihuai Gao, Dorsa Sadigh, and Shuran Song. 2025{\natexlab{c}}.
\newblock Unified video action model.
\newblock \emph{arXiv preprint arXiv:2503.00200}.

\bibitem[{Li et~al.(2024{\natexlab{b}})Li, Li, Liu, Wang, Liu, Kang, Ma, Kong,
  Zhang, and Liu}]{liu2024towards}
Xinghang Li, Peiyan Li, Minghuan Liu, Dong Wang, Jirong Liu, Bingyi Kang, Xiao
  Ma, Tao Kong, Hanbo Zhang, and Huaping Liu. 2024{\natexlab{b}}.
\newblock Towards generalist robot policies: What matters in building
  vision-language-action models.
\newblock \emph{arXiv preprint arXiv:2412.14058}.

\bibitem[{Lin et~al.(2025)Lin, Ding, Wang, Zhuang, Liu, Tong, Song, Lyu, Huang,
  and Wang}]{lin2025hifvla}
Minghui Lin, Pengxiang Ding, Shu Wang, Zifeng Zhuang, Yang Liu, Xinyang Tong,
  Wenxuan Song, Shangke Lyu, Siteng Huang, and Donglin Wang. 2025.
\newblock Hif-vla: Hindsight, insight and foresight through motion
  representation for vision-language-action models.
\newblock \emph{arXiv preprint arXiv:2512.09928}.

\bibitem[{Lipman et~al.(2023)Lipman, Chen, Ben-Hamu, Nickel, and
  Le}]{lipman2023flow}
Yaron Lipman, Ricky~TQ Chen, Heli Ben-Hamu, Maximilian Nickel, and Matthew Le.
  2023.
\newblock Flow matching for generative modeling.
\newblock In \emph{The Eleventh International Conference on Learning
  Representations}.

\bibitem[{Liu et~al.(2023)Liu, Zhu, Gao, Feng, Liu, Zhu, and
  Stone}]{liu2023libero}
Bo~Liu, Yifeng Zhu, Chongkai Gao, Yihao Feng, Qiang Liu, Yuke Zhu, and Peter
  Stone. 2023.
\newblock Libero: Benchmarking knowledge transfer for lifelong robot learning.
\newblock \emph{Advances in Neural Information Processing Systems},
  36:44776--44791.

\bibitem[{Mees et~al.(2022)Mees, Hermann, Rosete-Beas, and
  Burgard}]{mees2022calvin}
Oier Mees, Lukas Hermann, Erick Rosete-Beas, and Wolfram Burgard. 2022.
\newblock Calvin: A benchmark for language-conditioned policy learning for
  long-horizon robot manipulation tasks.
\newblock \emph{IEEE Robotics and Automation Letters}, 7(3):7327--7334.

\bibitem[{Mu et~al.(2025)Mu, Chen, Chen, Peng, Lan, Gao, Liang, Yu, Zou, Xu
  et~al.}]{mu2025robotwin}
Yao Mu, Tianxing Chen, Zanxin Chen, Shijia Peng, Zhiqian Lan, Zeyu Gao, Zhixuan
  Liang, Qiaojun Yu, Yude Zou, Mingkun Xu, and 1 others. 2025.
\newblock Robotwin: Dual-arm robot benchmark with generative digital twins.
\newblock In \emph{Proceedings of the Computer Vision and Pattern Recognition
  Conference}, pages 27649--27660.

\bibitem[{Prasad et~al.(2024)Prasad, Lin, Wu, Zhou, and
  Bohg}]{prasad2024consistency}
Aaditya Prasad, Kevin Lin, Jimmy Wu, Linqi Zhou, and Jeannette Bohg. 2024.
\newblock Consistency policy: Accelerated visuomotor policies via consistency
  distillation.
\newblock In \emph{Robotics: Science and Systems}.

\bibitem[{Salimans and Ho(2022)}]{salimans2022progressive}
Tim Salimans and Jonathan Ho. 2022.
\newblock Progressive distillation for fast sampling of diffusion models.
\newblock \emph{arXiv preprint arXiv:2202.00512}.

\bibitem[{Shi et~al.(2025)Shi, Xie, Liu, Sun, Liu, Wang, Zhou, Fan, Zhang, and
  Huang}]{shi2025memoryvla}
Hao Shi, Bin Xie, Yingfei Liu, Lin Sun, Fengrong Liu, Tiancai Wang, Erjin Zhou,
  Haoqiang Fan, Xiangyu Zhang, and Gao Huang. 2025.
\newblock Memoryvla: Perceptual-cognitive memory in vision-language-action
  models for robotic manipulation.
\newblock \emph{arXiv preprint arXiv:2508.19236}.

\bibitem[{Song et~al.(2021)Song, Meng, and Ermon}]{songdenoising}
Jiaming Song, Chenlin Meng, and Stefano Ermon. 2021.
\newblock Denoising diffusion implicit models.
\newblock In \emph{International Conference on Learning Representations}.

\bibitem[{Song et~al.(2023)Song, Dhariwal, Chen, and
  Sutskever}]{song2023consistency}
Yang Song, Prafulla Dhariwal, Mark Chen, and Ilya Sutskever. 2023.
\newblock Consistency models.
\newblock In \emph{International Conference on Machine Learning}, pages
  32211--32252. PMLR.

\bibitem[{Tian et~al.(2025)Tian, Yang, Zeng, Wang, Lin, Dong, and
  Pang}]{tianpredictive}
Yang Tian, Sizhe Yang, Jia Zeng, Ping Wang, Dahua Lin, Hao Dong, and Jiangmiao
  Pang. 2025.
\newblock Predictive inverse dynamics models are scalable learners for robotic
  manipulation.
\newblock In \emph{The Thirteenth International Conference on Learning
  Representations}.

\bibitem[{Wen et~al.(2025)Wen, Zhu, Zhu, Tang, Li, Zhou, Liu, Shen, Peng, and
  Feng}]{wen2025diffusionvla}
Junjie Wen, Yichen Zhu, Minjie Zhu, Zhibin Tang, Jinming Li, Zhongyi Zhou,
  Xiaoyu Liu, Chaomin Shen, Yaxin Peng, and Feifei Feng. 2025.
\newblock Diffusionvla: Scaling robot foundation models via unified diffusion
  and autoregression.
\newblock In \emph{Forty-second International Conference on Machine Learning}.

\bibitem[{Wen et~al.(2024)Wen, Lin, Zhu, Han, Xu, Zhao, and
  Liang}]{wen2024vidman}
Youpeng Wen, Junfan Lin, Yi~Zhu, Jianhua Han, Hang Xu, Shen Zhao, and Xiaodan
  Liang. 2024.
\newblock Vidman: Exploiting implicit dynamics from video diffusion model for
  effective robot manipulation.
\newblock \emph{Advances in Neural Information Processing Systems},
  37:41051--41075.

\bibitem[{Wu et~al.(2024)Wu, Jing, Cheang, Chen, Xu, Li, Liu, Li, and
  Kong}]{wuunleashing}
Hongtao Wu, Ya~Jing, Chilam Cheang, Guangzeng Chen, Jiafeng Xu, Xinghang Li,
  Minghuan Liu, Hang Li, and Tao Kong. 2024.
\newblock Unleashing large-scale video generative pre-training for visual robot
  manipulation.
\newblock In \emph{The Twelfth International Conference on Learning
  Representations}.

\bibitem[{Ze et~al.(2024)Ze, Zhang, Zhang, Hu, Wang, and Xu}]{ze20243d}
Yanjie Ze, Gu~Zhang, Kangning Zhang, Chenyuan Hu, Muhan Wang, and Huazhe Xu.
  2024.
\newblock 3d diffusion policy: Generalizable visuomotor policy learning via
  simple 3d representations.
\newblock In \emph{ICRA 2024 Workshop on 3D Visual Representations for Robot
  Manipulation}.

\bibitem[{Zhang et~al.(2025{\natexlab{a}})Zhang, Guo, Hu, Chen, Zhu, and
  Chen}]{zhang2025up}
Jianke Zhang, Yanjiang Guo, Yucheng Hu, Xiaoyu Chen, Xiang Zhu, and Jianyu
  Chen. 2025{\natexlab{a}}.
\newblock Up-vla: A unified understanding and prediction model for embodied
  agent.
\newblock In \emph{International Conference on Machine Learning}, pages
  74911--74922. PMLR.

\bibitem[{Zhang et~al.(2025{\natexlab{b}})Zhang, Sha, Jiang, Loper, Song, Cai,
  Xu, Hu, Zheng, and Li}]{zhang2025real}
Kaifeng Zhang, Shuo Sha, Hanxiao Jiang, Matthew Loper, Hyunjong Song, Guangyan
  Cai, Zhuo Xu, Xiaochen Hu, Changxi Zheng, and Yunzhu Li. 2025{\natexlab{b}}.
\newblock Real-to-sim robot policy evaluation with gaussian splatting
  simulation of soft-body interactions.
\newblock \emph{arXiv preprint arXiv:2511.04665}.

\bibitem[{Zhang et~al.(2025{\natexlab{c}})Zhang, Liu, Qi, Wang, Yu, Zhang,
  Dong, He, Wang, Zhang et~al.}]{zhangdreamvla}
Wenyao Zhang, Hongsi Liu, Zekun Qi, Yunnan Wang, XinQiang Yu, Jiazhao Zhang,
  Runpei Dong, Jiawei He, He~Wang, Zhizheng Zhang, and 1 others.
  2025{\natexlab{c}}.
\newblock Dreamvla: A vision-language-action model dreamed with comprehensive
  world knowledge.
\newblock In \emph{The Thirty-ninth Annual Conference on Neural Information
  Processing Systems}.

\bibitem[{Zhao et~al.(2025)Zhao, Lu, Kim, Fu, Zhang, Wu, Li, Ma, Han, Finn
  et~al.}]{zhao2025cot}
Qingqing Zhao, Yao Lu, Moo~Jin Kim, Zipeng Fu, Zhuoyang Zhang, Yecheng Wu,
  Zhaoshuo Li, Qianli Ma, Song Han, Chelsea Finn, and 1 others. 2025.
\newblock Cot-vla: Visual chain-of-thought reasoning for vision-language-action
  models.
\newblock In \emph{Proceedings of the Computer Vision and Pattern Recognition
  Conference}, pages 1702--1713.

\bibitem[{Zhao et~al.(2023)Zhao, Kumar, Levine, and Finn}]{zhao2023learning}
Tony Zhao, Vikash Kumar, Sergey Levine, and Chelsea Finn. 2023.
\newblock Learning fine-grained bimanual manipulation with low-cost hardware.
\newblock \emph{Robotics: Science and Systems XIX}.

\bibitem[{Zheng et~al.(2025)Zheng, Liang, Huang, Gao, Daum{\'e}~III, Kolobov,
  Huang, and Yang}]{zhengtracevla}
Ruijie Zheng, Yongyuan Liang, Shuaiyi Huang, Jianfeng Gao, Hal Daum{\'e}~III,
  Andrey Kolobov, Furong Huang, and Jianwei Yang. 2025.
\newblock Tracevla: Visual trace prompting enhances spatial-temporal awareness
  for generalist robotic policies.
\newblock In \emph{The Thirteenth International Conference on Learning
  Representations}.

\bibitem[{Zhong et~al.(2025)Zhong, Yan, Li, Liu, Gong, Zhang, Song, Chen,
  Zheng, Wang et~al.}]{zhong2025flowvla}
Zhide Zhong, Haodong Yan, Junfeng Li, Xiangchen Liu, Xin Gong, Tianran Zhang,
  Wenxuan Song, Jiayi Chen, Xinhu Zheng, Hesheng Wang, and 1 others. 2025.
\newblock Flowvla: Visual chain of thought-based motion reasoning for
  vision-language-action models.
\newblock \emph{arXiv preprint arXiv:2508.18269}.

\bibitem[{Zitkovich et~al.(2023)Zitkovich, Yu, Xu, Xu, Xiao, Xia, Wu, Wohlhart,
  Welker, Wahid et~al.}]{brohan2023rt2}
Brianna Zitkovich, Tianhe Yu, Sichun Xu, Peng Xu, Ted Xiao, Fei Xia, Jialin Wu,
  Paul Wohlhart, Stefan Welker, Ayzaan Wahid, and 1 others. 2023.
\newblock Rt-2: Vision-language-action models transfer web knowledge to robotic
  control.
\newblock In \emph{Conference on Robot Learning}, pages 2165--2183. PMLR.

\end{thebibliography}

\appendix

\section{Use of AI Assistants}
\label{app:ai_use}

AI assistants were used solely for grammar checking and polishing.

\section{Qualitative Visualization on Simulation Benchmarks}
\label{app:simulation_viz}

To further supplement the quantitative experimental analysis, this section provides detailed qualitative visualization results of ElasticFlow in two high-fidelity simulation environments: \textbf{RoboCasa} and \textbf{RoboTwin}.

\begin{figure*}[t] 
	\centering
	\includegraphics[width=\linewidth]{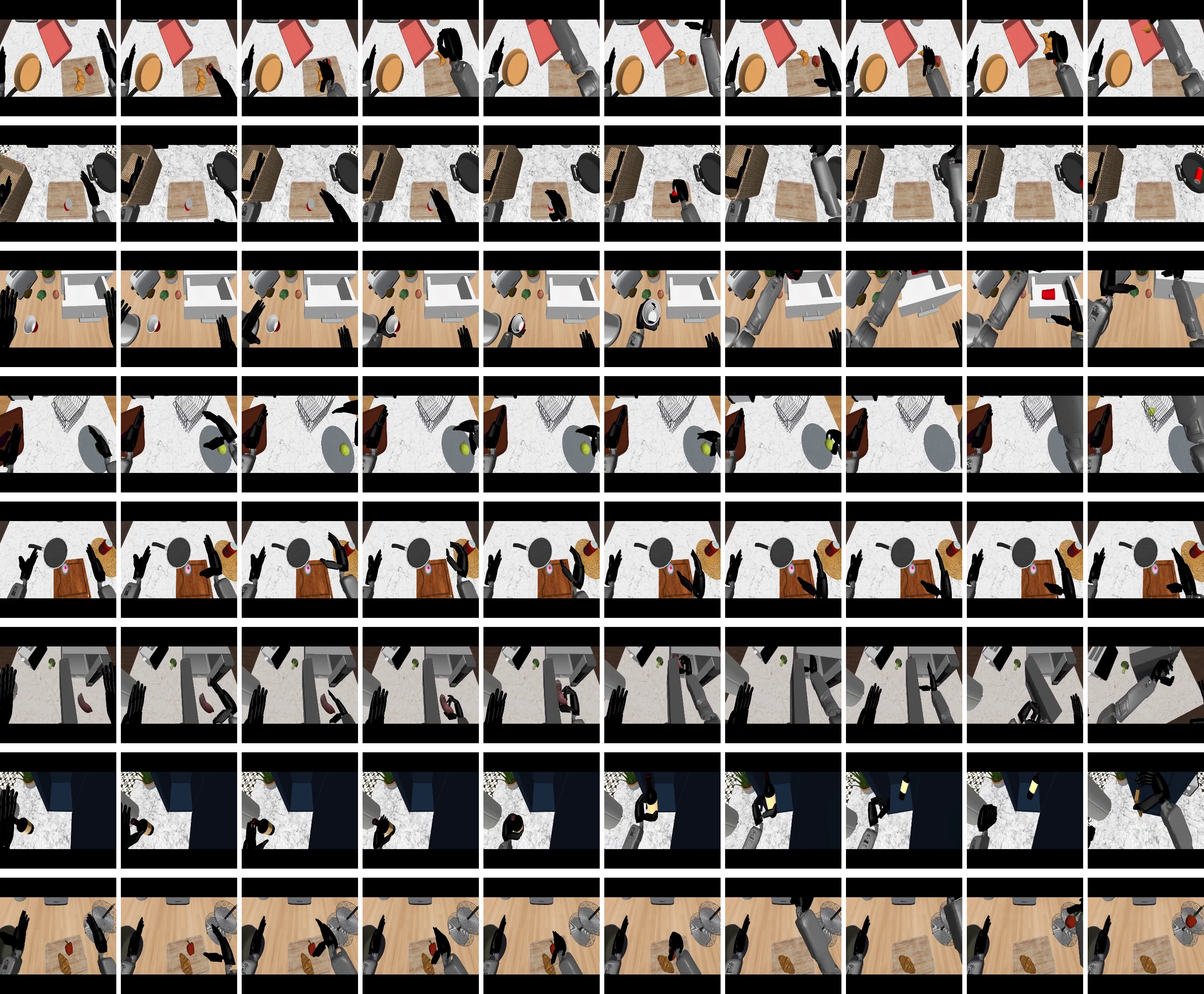} 
	\caption{\textbf{Qualitative Visualization of ElasticFlow on RoboCasa Benchmark.} Each row in the figure displays a complete kitchen manipulation task sequence (e.g., food preparation, cabinet interaction). Thanks to the Elastic Time Horizon mechanism, the model exhibits excellent temporal consistency and action smoothness in these long-horizon tasks involving multi-stage planning.}
	\label{fig:robocasa_viz}
\end{figure*}

\paragraph{RoboCasa Visualization} As shown in Figure~\ref{fig:robocasa_viz}, we display generation trajectories for tasks involving complex contact dynamics and multi-stage logic (e.g., opening a microwave). These results intuitively verify the ability of the Elastic Time Horizon mechanism to maintain temporal consistency and action smoothness, effectively avoiding cumulative errors in long-duration operations.

\begin{figure*}[t]
	\centering
	\includegraphics[width=\linewidth]{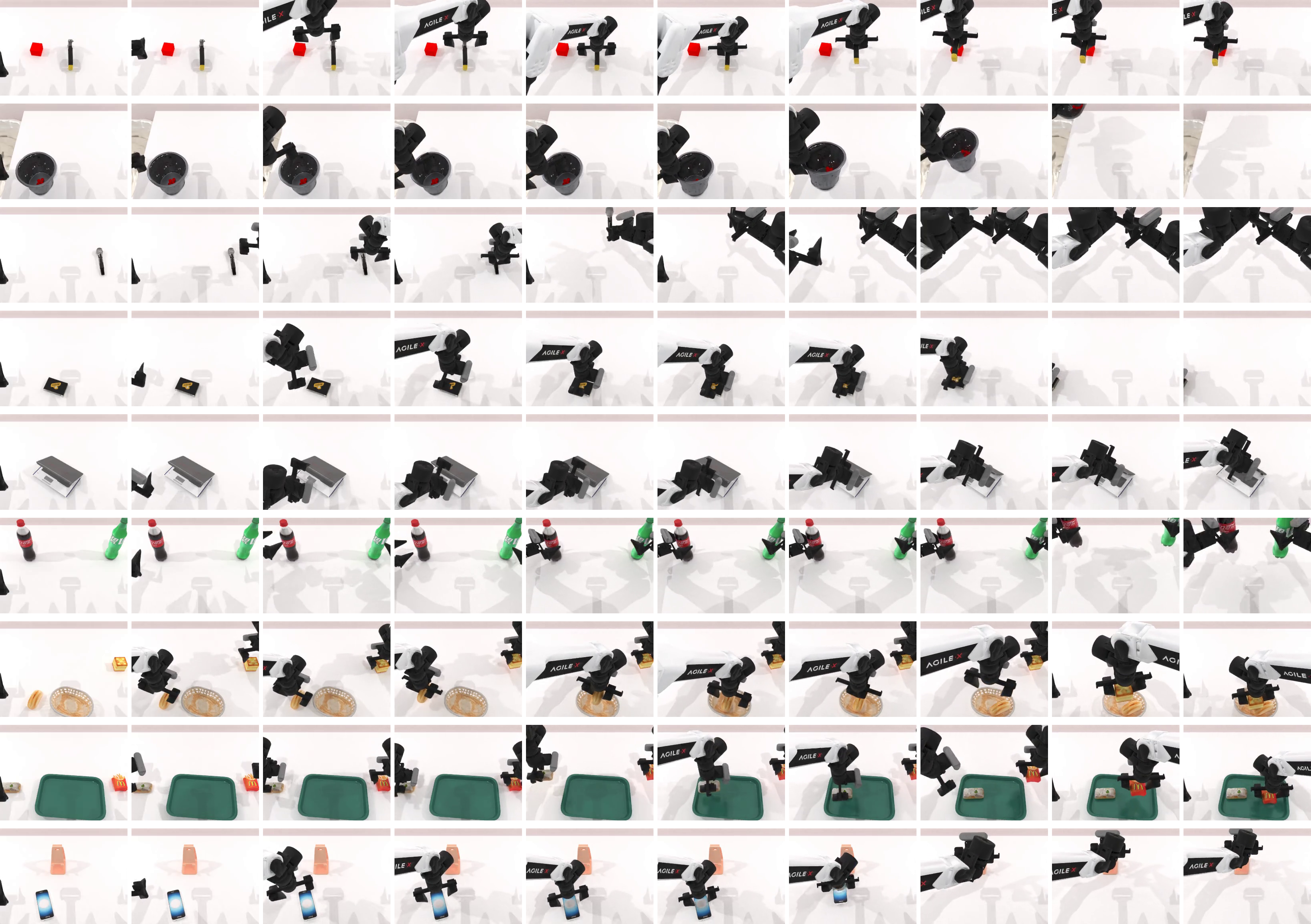}
	\vspace{-10pt} 
	\caption{\textbf{Qualitative Visualization of ElasticFlow in RoboTwin2.0 Benchmark.} Each row displays a continuous execution process of a different task, covering various operation scenarios ranging from short horizons (e.g., lifting blocks) to long horizons (e.g., object switching, organizing). As shown, thanks to the Elastic Time Horizon mechanism, ElasticFlow generates smooth, physically consistent, and temporally coherent action sequences without significant error accumulation in long-range tasks.}
	\label{fig:robotwin_viz}
	\vspace{-10pt} 
\end{figure*}

\paragraph{RoboTwin Visualization} Figure~\ref{fig:robotwin_viz} displays continuous execution processes across varying task horizons, ranging from short (e.g., lifting blocks) to long (e.g., object organizing). As illustrated, ElasticFlow generates smooth, physically consistent, and temporally coherent action sequences without significant error accumulation, further validating the effectiveness of our proposed mechanism across diverse task horizons.

\section{ElasticFlow Mathematical Derivation and Proof}
\label{app:derivation}

This section provides the detailed mathematical derivation process for the ElasticFlow Identity (Equation \ref{eq:ElasticFlow_identity}) in the main text, as well as a detailed explanation regarding the calculation of the total derivative.

\subsection{Derivation of ElasticFlow Identity}

Our goal is to establish a differential connection between the average velocity field $u(z_t, r, t)$ and the instantaneous velocity field $v(z_t, t)$. Recall the definition of the average velocity field:
\begin{equation}
	u(z_t, r, t) \triangleq \frac{1}{t-r} \int_{r}^{t} v(z_\tau, \tau) d\tau
\end{equation}
Rewrite it in integral form to eliminate the time term in the denominator:
\begin{equation}
	(t-r) u(z_t, r, t) = \int_{r}^{t} v(z_\tau, \tau) d\tau
	\label{eq:app_integral}
\end{equation}

To construct a supervision signal dependent only on the current state $z_t$, we take the Total Derivative with respect to time $t$ on both sides of Equation (\ref{eq:app_integral}).

For the left side, using the Product Rule:
\begin{equation}
	\frac{d}{dt} \left[ (t-r) u(z_t, r, t) \right] = 1 \cdot u(z_t, r, t) + (t-r) \frac{d}{dt} u(z_t, r, t)
	\label{eq:app_lhs}
\end{equation}

For the right side, according to the Fundamental Theorem of Calculus, the derivative of an integral function with respect to its upper limit is the value of the integrand at that limit. Note that here $v(z_\tau, \tau)$ is defined along the streamline, so the total derivative with respect to $t$ acts directly on the integral upper limit:
\begin{equation}
	\frac{d}{dt} \int_{r}^{t} v(z_\tau, \tau) d\tau = v(z_t, t)
	\label{eq:app_rhs}
\end{equation}

Combining equations (\ref{eq:app_lhs}) and (\ref{eq:app_rhs}), we get:
\begin{equation}
	u(z_t, r, t) + (t-r) \frac{d}{dt} u(z_t, r, t) = v(z_t, t)
\end{equation}

Rearranging terms yields the ElasticFlow Identity presented in the main text:
\begin{equation}
	u(z_t, r, t) = v(z_t, t) - (t-r) \frac{d}{dt} u(z_t, r, t)
\end{equation}
Q.E.D.

\subsection{Expansion of Total Derivative and JVP Calculation}

In the above identity, $\frac{d}{dt} u(z_t, r, t)$ is the total derivative. Since $u$ explicitly depends on time $t$ and implicitly depends on $t$ through state $z_t$, we need to expand it using the Chain Rule for multivariate functions:
\begin{equation}
	\frac{d}{dt} u(z_t, r, t) = \underbrace{\nabla_{z} u(z_t, r, t) \cdot \frac{dz_t}{dt}}_{\text{Convective Term}} + \underbrace{\frac{\partial}{\partial t} u(z_t, r, t)}_{\text{Unsteady Term}}
\end{equation}

According to the definition of the flow model, the rate of change of the state with respect to time is the instantaneous velocity, i.e., $\frac{dz_t}{dt} = v(z_t, t)$. Substituting this into the above equation yields:
\begin{equation}
	\frac{d}{dt} u(z_t, r, t) = \nabla_{z} u(z_t, r, t) \cdot v(z_t, t) + \frac{\partial}{\partial t} u(z_t, r, t)
\end{equation}

In actual neural network training, directly computing the Jacobian matrix $\nabla_{z} u$ (dimension $D \times D$) is computationally too expensive. We utilize Forward-mode Automatic Differentiation to directly calculate the Jacobian-Vector Product (JVP):
\begin{equation}
	\text{JVP}(u_\theta, v) = \nabla_{z} u_\theta \cdot v
\end{equation}
This makes the complexity of computing the total derivative comparable to a single backpropagation pass, ensuring training efficiency.

\subsection{Sensitivity Analysis of CFG Guidance Weight}
\label{app:cfg_ablation}

The Classifier-Free Guidance (CFG) weight $w$ is a key hyperparameter balancing generation diversity and conditional consistency. We tested the impact of $w \in [1.0, 4.0]$ on task success rates on the RoboTwin Long Horizon task subset.

\begin{figure}[t] 
	\centering
	\includegraphics[width=1.0\linewidth]{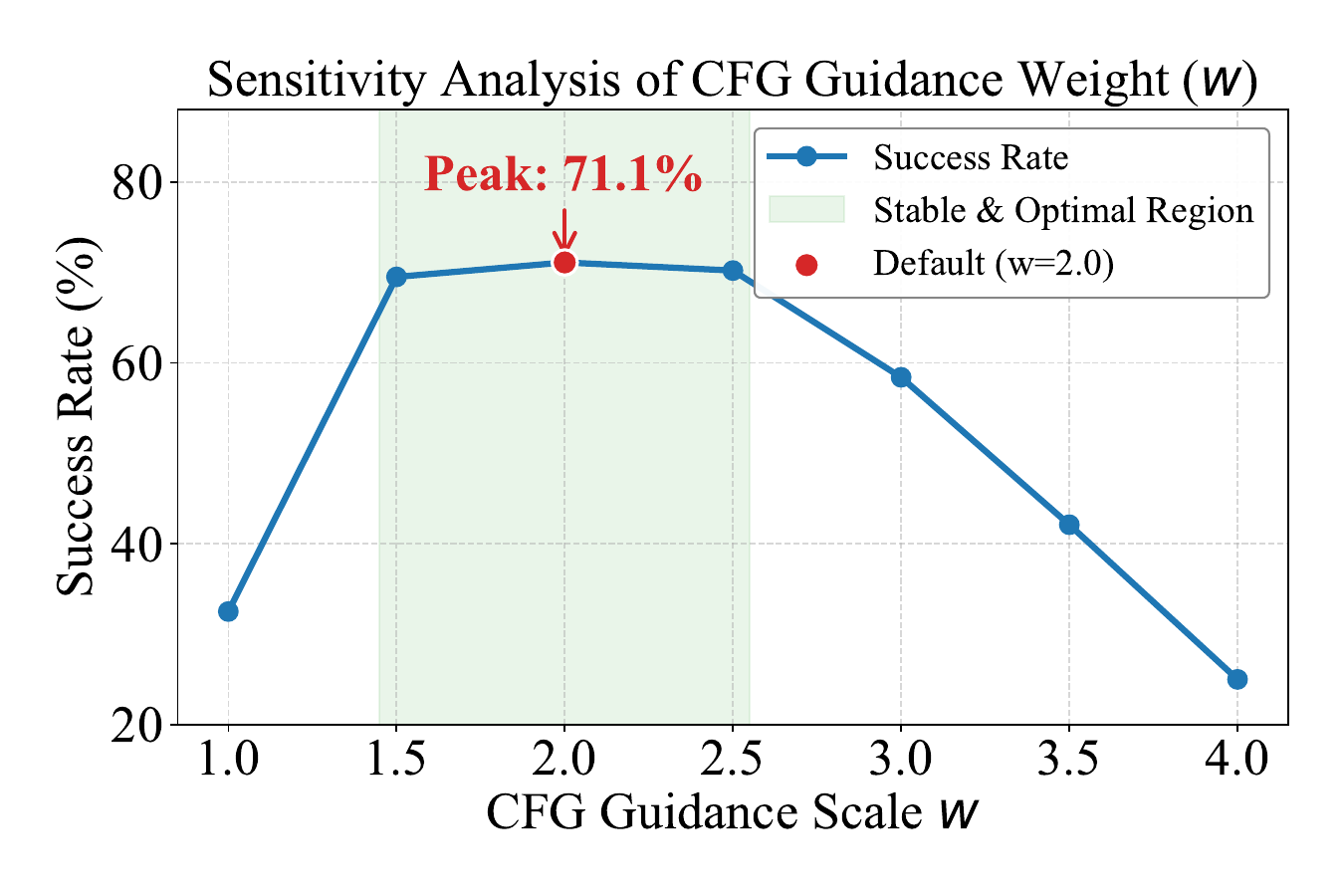}
	\caption{\textbf{CFG Weight Sensitivity Analysis.} The curve shows the trend of ElasticFlow's success rate on RoboTwin long-horizon tasks as $w$ changes. The \textbf{Red Node} ($w=2.0$) marks the peak success rate (71.1\%); the \textbf{Green Region} indicates the optimal parameter interval where model performance is robust ($w \in [1.5, 2.5]$).}
	\label{fig:cfg_sensitivity}
\end{figure}

The experimental results are shown in Figure~\ref{fig:cfg_sensitivity}. We observe:
\begin{itemize}
	\item \textbf{Insufficient Guidance ($w=1.0$)}: The model's ability to follow fine-grained language instructions is weak, resulting in a success rate of only 32.5\%.
	\item \textbf{Optimal Interval ($w \in [1.5, 2.5]$)}: Performance improves significantly as $w$ increases. It peaks at \textbf{71.1\%} when $w=2.0$ and remains highly stable within the $[1.5, 2.5]$ range.
	\item \textbf{Excessive Guidance ($w > 3.0$)}: overly strong guidance leads to rigid action trajectories and high-frequency jitter (Oversaturation), causing a sharp drop in success rate.
\end{itemize}

Based on this analysis, we set $w=2.0$ as the default for all major experiments.

\section{Hardware Configuration Details}
\label{app:hardware}

We list the detailed hardware specifications used for model training and inference evaluation in Table~\ref{tab:hardware_specs}. Considering the high demand for VRAM bandwidth during training, we conducted model training on high-performance server nodes equipped with NVIDIA A100 GPUs. To demonstrate ElasticFlow's operational efficiency and deployability in real-world scenarios, all inference latency tests and real-robot experiments were completed on standard workstations equipped with consumer-grade NVIDIA RTX 4090 GPUs.

\begin{table*}[h]
	\centering
	\caption{\textbf{Detailed Hardware Configuration List.} This study used high-performance computing nodes for training and standard workstations for inference evaluation.}
	\label{tab:hardware_specs}
	\vspace{0.2cm}
	\small
	\renewcommand{\arraystretch}{1.2}
	\begin{tabular}{l|p{6cm}|p{6cm}}
		\toprule
		\textbf{Component} & \textbf{Training Environment} & \textbf{Inference Environment} \\
		\midrule
		\textbf{GPU} & 4 $\times$ NVIDIA A100-SXM4 (80GB VRAM) & 1 $\times$ NVIDIA GeForce RTX 4090 (24GB VRAM) \\
		\textbf{CPU} & Intel Xeon Platinum 8358P @ 2.60GHz & Intel Xeon Gold 6430 \\
		\textbf{RAM} & 1.0 TB & 503 GB \\
		\textbf{OS} & Ubuntu 22.04.3 LTS & Ubuntu 22.04.5 LTS \\
		\textbf{CUDA Version} & 12.4 & 12.9 \\
		\bottomrule
	\end{tabular}
\end{table*}

\section{Real-World Experiment Details}
\label{app:real_world_details}

This section provides the detailed setup, task definitions, and extensive quantitative evaluation results for the real-world experiments.

\subsection{Hardware Setup \& Protocol}
All real-world experiments were conducted on a UFACTORY xArm6 robotic arm equipped with a Robotiq 2F-85 parallel gripper. Visual input was provided by two Intel RealSense D435i cameras, mounted in third-person view (global) and wrist view (local), respectively. Inference was performed on a single NVIDIA RTX 4090 GPU, with control frequency stable at \textbf{71Hz} and end-to-end latency below 15ms.

To comprehensively evaluate ElasticFlow's robustness, we divided test conditions into two categories:
\begin{itemize}
	\item \textbf{Seen Scenarios (In-Distribution):} Object positions, lighting conditions, and distractor distribution remained consistent with data collection (20-50 trajectories per task).
	\item \textbf{Unseen Scenarios (Out-of-Distribution/Generalization):} Variables not present in training data were introduced, including: (1) \textbf{Spatial Generalization} (object position shift $\pm 10$cm, rotation $\pm 30^\circ$); (2) \textbf{Instance Generalization} (using new objects with different shapes or colors); (3) \textbf{Dynamic Disturbance} (manually moving the target object at $\sim 5$cm/s during execution).
\end{itemize}

\subsection{Task Definitions and Challenges}
We designed 7 tasks with different physical characteristics to fully verify ElasticFlow's advantages in high-frequency response, contact-rich manipulation, and long-range planning. Table~\ref{tab:real_world_tasks} details the correspondence between each task and the core contributions of this paper (physical consistency, elastic horizon).

\begin{table*}[h]
	\centering
	\caption{\textbf{Real-World Task Definitions and Physical Challenges.} Tasks range from short-horizon high-frequency reactions to long-horizon structured planning.}
	\label{tab:real_world_tasks}
	\small
	\renewcommand{\arraystretch}{1.1} 
	\begin{tabular}{p{3.2cm} p{2.2cm} p{9.0cm}}
		\toprule
		\textbf{Task} & \textbf{Horizon} & \textbf{Key Challenges and Physical Characteristics} \\
		\midrule
		\textbf{1. Dynamic Interception} & Short / Reactive & \textbf{High-Frequency Response.} The robot must intercept a cylinder rolling at random speeds on a table. Tests the real-time tracking capability of the 71Hz control loop for moving targets. \\
		\midrule 
		\textbf{2. Precision Insertion} & Short / Contact & \textbf{Jitter-Free.} Inserting a deformable straw or metal pin into a tight-fitting holder. Tests MeanFlow's ability to eliminate high-frequency end-effector jitter and prevent object damage. \\
		\midrule
		\textbf{3. Liquid Pouring} & Short / Smoothness & \textbf{Trajectory Smoothness.} Pouring a water-filled cup into another container without spilling. Tests consistency of the generated trajectory in terms of velocity and acceleration (Low Jerk). \\
		\midrule
		\textbf{4. Cable Routing} & Medium / Deformable & \textbf{Non-Rigid Dynamics.} Routing a soft cable around obstacles and arranging it into a specific shape. Tests the model's perception and prediction of deformable object states. \\
		\midrule
		\textbf{5. Unstable Stacking} & Medium / Stability & \textbf{Contact Stability.} Stacking objects with irregular shapes or low friction (e.g., markers). Any minor generation error can cause the stack to collapse. \\
		\midrule
		\textbf{6. Tool Use \& Hammering} & Medium / Tool Use & \textbf{End-Effector Extension.} Grasping a hammer and accurately striking a target nail. Tests the model's adaptability to kinematic changes of the end-effector and contact force control. \\
		\midrule
		\textbf{7. Long-Horizon Kitchen} & Long / Sequential & \textbf{Temporal Consistency.} Continuously executing "Open microwave $\to$ Put in bowl $\to$ Close door $\to$ Press switch". Tests the ability of Elastic Time Horizon to maintain global structure in multi-stage tasks. \\
		\bottomrule
	\end{tabular}
\end{table*}

\subsection{Quantitative Results}
We conducted 20 real-machine trials for each task under both Seen and Unseen settings (totaling $7 \times 2 \times 20 = 280$ trials).

As shown in Table~\ref{tab:real_world_comprehensive}, ElasticFlow achieved an average success rate of 91.4\% in Seen scenarios. More importantly, in Unseen scenarios, despite facing unseen object instances or dynamic disturbances, the model maintained an average success rate of 76.4\%. Particularly in the "Dynamic Interception" task, thanks to the low latency of one-step inference, the model effectively intercepted rolling targets with speeds up to $\sim 10$cm/s, demonstrating Sim-to-Real robustness significantly superior to traditional low-frequency policies.

\begin{table*}[h]
	\centering
	\caption{\textbf{Quantitative Evaluation Results of Real-World Experiments (Success Rate \%).} We compare model performance within training distribution (Seen) and out-of-distribution (Unseen). Unseen settings include New Instance, Position Shift, or Dynamic Disturbance.}
	\label{tab:real_world_comprehensive}
	\vspace{0.2cm}
	\resizebox{0.8\linewidth}{!}{
		\begin{tabular}{lccc}
			\toprule
			\multirow{2}{*}{\textbf{Task}} & \textbf{Seen Scenario} & \multicolumn{2}{c}{\textbf{Unseen Scenario (Generalization)}} \\
			\cmidrule(lr){2-2} \cmidrule(lr){3-4}
			& \textbf{Success Rate} & \textbf{Variation Type} & \textbf{Success Rate} \\
			\midrule
			1. Dynamic Interception & 95\% (19/20) & Variable Speed ($\le 10$cm/s) & 85\% (17/20) \\
			2. Precision Insertion & 90\% (18/20) & Position Shift ($\pm 5$cm) & 80\% (16/20) \\
			3. Liquid Pouring & 95\% (19/20) & New Cup Instance (Color) & 85\% (17/20) \\
			4. Cable Routing & 85\% (17/20) & Stiffer Cable Material & 70\% (14/20) \\
			5. Unstable Stacking & 85\% (17/20) & New Object Geometry & 65\% (13/20) \\
			6. Tool Use \& Hammering & 90\% (18/20) & Distractor Objects Added & 75\% (15/20) \\
			7. Long-Horizon Kitchen & 100\% (20/20) & Start Position Shift & 75\% (15/20) \\
			\midrule
			\textbf{Average} & \textbf{91.4\%} & - & \textbf{76.4\%} \\
			\bottomrule
		\end{tabular}
	}
\end{table*}

\section{Additional Experimental Results}
\label{app:additional_experiments}

In this section, we provide detailed quantitative breakdowns for the RoboTwin and LIBERO benchmarks to supplement the main text.

\subsection{RoboTwin Horizon Analysis}
Table~\ref{tab:robotwin2_results_by_difficulty} presents the comprehensive results on the \textbf{RoboTwin2.0} benchmark, categorized by task horizon difficulty. 
These results quantitatively verify that ElasticFlow maintains superior stability in long-horizon tasks compared to baselines like $\pi_0$ and RDT.

\begin{table*}[!t]
	\centering
	\caption{Main results on \textbf{RoboTwin2.0}, organized by task horizon difficulty. Our method demonstrates superior stability in long-horizon tasks.}
	\resizebox{\linewidth}{!}{
		\begin{tabular}{lccccc}
			\toprule
			\multicolumn{6}{c}{\textbf{Short Horizon Tasks (100-130 Steps)}} \\
			\midrule
			\textbf{Model} & \textbf{Lift Pot} & \textbf{Beat Hammer Block} & \textbf{Pick Dual Bottles} & \textbf{Place Phone Stand} & \textbf{Avg} \\
			\midrule
			$\pi_0$ & 51.0 & 59.0 & 50.0 & 22.0 & 45.5 \\
			RDT & 45.0 & 22.0 & 18.0 & 13.0 &24.5 \\
			\midrule
			OpenVLA-OFT & 10.1 & 28.1 & 29.7 & 17.1 & 21.3 \\
			SimpleVLA-RL\citep{li2025simplevla} & 64.1 & \textbf{87.5} & 68.3 & 39.6 & 64.9 \\
			\textbf{ElasticFlow (Ours)} & \textbf{67.4} & 83.8 & \textbf{69.2} & \textbf{41.3} & \textbf{65.4} \\
			\midrule
			\multicolumn{6}{c}{\textbf{Medium Horizon Tasks (150-230 Steps)}} \\
			\midrule
			\textbf{Model} & \textbf{Move Can Pot} & \textbf{Place A2B Left} & \textbf{Place Empty Cup} & \textbf{Handover Mic} & \textbf{Avg} \\
			\midrule
			$\pi_0$ & 41.0 & 38.0 & 60.0 & 96.0 & 58.8 \\
			RDT & 33.0 & 21.0 & 42.0 & 95.0 &47.8 \\
			\midrule
			OpenVLA-OFT & 28.1 & 37.5 & 77.3 & 45.3 & 47.1 \\
			SimpleVLA-RL\citep{li2025simplevla} & 61.2 & 45.3 & \textbf{94.2} & \textbf{89.2} & 72.5 \\
			\textbf{ElasticFlow (Ours)} & \textbf{63.4} & \textbf{51.7} & 92.1 & 87.6 & \textbf{73.7} \\
			\midrule
			\multicolumn{6}{c}{\textbf{Long (280-320 Steps) \& Extra Long Horizon Tasks (450-650 Steps)}} \\
			\midrule
			\textbf{Model} & \textbf{Handover Block} & \textbf{Stack Bowls Two} & \textbf{Blocks Rank Rgb} & \textbf{Put Bottles Dustbin} & \textbf{Avg} \\
			\midrule
			$\pi_0$ & 39.0 & 53.0 & 45.0 & 36.0 & 43.3 \\
			RDT & 26.0 & 42.0 & 17.0 & 26.0 & 27.8 \\
			\midrule
			OpenVLA-OFT & 33.1 & 40.6 & 70.2 & 42.2 & 46.5 \\
			SimpleVLA-RL\citep{li2025simplevla} & 57.8 & 75.8 & \textbf{81.3} & 60.9 & 69.0 \\
			\textbf{ElasticFlow (Ours)} & \textbf{62.9} & \textbf{78.3} & 79.8 & \textbf{63.4} & \textbf{71.1} \\
			\midrule
			\textbf{Overall Avg} & \multicolumn{5}{l}{\footnotesize \textbf{RDT:} 33.3 \quad $\mathbf{\pi_0}$: 49.2 \quad \textbf{OpenVLA-OFT:} 38.3 \quad \textbf{SimpleVLA:} 68.8 \quad \textbf{Ours: 70.1}} \\
			\bottomrule
		\end{tabular}
	}
	\label{tab:robotwin2_results_by_difficulty}
\end{table*}

\subsection{LIBERO-Long Detailed Analysis}
We also provide the detailed success rate breakdown for the 10 long-horizon tasks in the \textit{LIBERO-Long} suite (see Table~\ref{tab:libero_long}).
This suite requires the robot to complete continuous sub-goals.
Results show that ElasticFlow achieved an average success rate of \textbf{95.6\%} under third-person perspective and \textbf{97.6\%} under multi-view input.
In comparison, OpenVLA (54.0\%), which relies on discrete Tokenized output, underperformed.
This suggests that our continuous flow method effectively avoids the quantization errors that typically accumulate in long horizons.
Furthermore, compared to MemoryVLA (93.4\%), ElasticFlow demonstrated stronger robustness in complex tasks such as "Put both pots on stove" (+19.0\% improvement), verifying that the Elastic Time Horizon mechanism helps maintain structural consistency of trajectory planning over extended operation periods.

\begin{table*}[ht]
	\caption{Performance comparison on \textbf{LIBERO-Long} benchmark. We report the success rate (\%) on 10 long-horizon tasks. Methods marked with ``$*$'' are reproduced using official implementations. \textbf{Bold} indicates the best performance.}
	\label{tab:libero_long}
	\centering
	\begingroup
	\small
	\setlength{\tabcolsep}{3pt}
	\renewcommand{\arraystretch}{1.1}

	\renewcommand{\tabularxcolumn}[1]{b{#1}}

	\begin{tabularx}{\linewidth}{@{} p{2.3cm} p{0.8cm} *{10}{Y} @{}}
		\toprule
		\textbf{Method} & 
		\textbf{\tiny \makecell[bc]{Avg.\\SR}} & 
		\tiny \makecell[bc]{Put soup\\\& box\\in basket} &
		\tiny \makecell[bc]{Put box\\\& butter\\in basket} &
		\tiny \makecell[bc]{Turn on\\stove \&\\put pot} &
		\tiny \makecell[bc]{Put bowl\\in drawer\\\& close} &
		\tiny \makecell[bc]{Put mugs\\on L\&R\\plates} &
		\tiny \makecell[bc]{Pick book\\\& place\\in back} &
		\tiny \makecell[bc]{Mug on\\plate,\\pudding R} &
		\tiny \makecell[bc]{Put soup\\\& sauce\\in basket} &
		\tiny \makecell[bc]{Put both\\pots on\\stove} &
		\tiny \makecell[bc]{Put mug \\in\\microwave\\and close} \\
		\midrule
		OpenVLA~\citep{kim2025openvla} & 54.0 & 35.0 & 95.0 & 65.0 & 45.0 & 40.0 & 80.0 & 60.0 & 45.0 & 20.0 & 55.0 \\
		UniVLA*~\citep{li2025unified} & 63.0 & 64.0 & 82.0 & 76.0 & 96.0 & 58.0 & {98.0} & 24.0 & 74.0 & 32.0 & 26.0 \\
		MemoryVLA~\citep{shi2025memoryvla} & 93.4 & 92.0 & {96.0} & 96.0 & \textbf{100} & \textbf{100} & \textbf{100} & \textbf{96.0} & 96.0 & 62.0 & \textbf{{96.0}} \\
		\footnotesize {OpenVLA-OFT}*~\citep{kim2025fine} & 91.0 & 82.0 & {96.0} & 96.0 & 94.0 & 90.0 & 96.0 & {92.0} & \textbf{100} & 70.0 & 94.0 \\

		HiF-VLA~\citep{lin2025hifvla} & 94.4 & 94.0 & \textbf{98.0} & \textbf{100} & \textbf{100} & 94.0 & \textbf{100} & 90.0 & {98.0} & 76.0 & 94.0 \\
		\rowcolor{gray!15}
		\textbf{ElasticFlow (Ours)} & \textbf{95.6} & \textbf{95.0} & 97.0 & 98.0 & 99.0 & 99.0 & 99.0 & 93.0 & \textbf{100.0} & \textbf{81.0} & 95.0
		\\
		\midrule
		\multicolumn{11}{l}{\textit{Policy inputs: third-person image, wrist-view image, language instruction}} \\
		\midrule
		Seer (scratch)~\citep{tianpredictive} & 78.7 & 80.0 & 90.0 & 91.7 & 81.7 & 85.0 & 65.0 & 86.7 & 88.3 & 51.7 & 66.7 \\
		Seer~\citep{tianpredictive} & 87.7 & 91.7 & 90.0 & 98.3 & \textbf{100} & 91.7 & 93.3 & 85.0 & 88.3 & 61.7 & 71.7 \\
		UniVLA*~\citep{li2025unified} & 90.0 & \textbf{100} & 92.0 & 94.0 & {98.0} & 86.0 & \textbf{100} & 80.0 & \textbf{100} & 70.0 & 82.0 \\
		\footnotesize {OpenVLA-OFT}*~\citep{kim2025fine} & 94.0 & 90.0 & \textbf{98.0} & {98.0} & {98.0} & {96.0} & \textbf{100} & {92.0} & \textbf{100} & 72.0 & {96.0} \\
		HiF-VLA~\citep{lin2025hifvla} & 96.4 & 88.0 & \textbf{98.0} & \textbf{100} & \textbf{100} & \textbf{100} & \textbf{100} & 96.0 & \textbf{100} & 82.0 & \textbf{100} \\
		\rowcolor{gray!15}
		\textbf{ElasticFlow (Ours)} & \textbf{97.6} & 93.0 & 97.0 & 99.0 & 99.0 & 99.0 & \textbf{100} & \textbf{100} & 98.0 & \textbf{91.0} & \textbf{100}
		\\
		\bottomrule
	\end{tabularx}
	\vspace{-8pt}
	\endgroup
\end{table*}

\end{document}